\RequirePackage{fix-cm}
\documentclass{preprint-svjour3}                     % onecolumn (standard format)
\smartqed  % flush right qed marks, e.g. at end of proof
\usepackage{graphicx}
\usepackage{subcaption}
\usepackage{amsmath}
\usepackage{multirow}
\usepackage{tabularx}
\usepackage{booktabs}
\usepackage{hyperref}
\usepackage{listings}
\usepackage{wrapfig}
\usepackage[numbers]{natbib}
\usepackage[section]{placeins}
\usepackage{hyperref}
\usepackage{doi}
\usepackage[table]{xcolor}
\begin{document}

\title{ Tag-based regulation of modules in genetic programming improves context-dependent problem solving }
% tag-based genetic regulation for event-driven genetic programs?

% \subtitle{Do you have a subtitle?\\ If so, write it here}

\titlerunning{Tag-based regulation of modules in genetic programming}

\author{
Alexander Lalejini    \and
Matthew Andres Moreno \and
Charles Ofria
}


% ORIGINAL:
% \institute{F. Author \at
%               first address \\
%               Tel.: +123-45-678910\\
%               Fax: +123-45-678910\\
%               \email{fauthor@example.com}           %  \\
% %             \emph{Present address:} of F. Author  %  if needed
%               second address
% }

\institute{
Alexander Lalejini$^{1,2,3}$ \\
\email{amlalejini@gmail.com} \\
Matthew Andres Moreno$^{1,2,3}$ \\
\email{mmore500@msu.edu} \\
Charles Ofria$^{1,2,3}$ \\
\email{ofria@msu.edu} \\
$^1$BEACON Center for the Study of Evolution in Action, Michigan State University \\
$^2$Department of Computer Science and Engineering, Michigan State University \\
$^3$Ecology, Evolution, and Behavior Program, Michigan State University
}

\date{\ }

\maketitle

%%%%%%%%%%%%%%%%%%%%%%%%%%%%%%%%%%%%%%%%%%%%%%%%%%%%%%%%%%%
% Utility variables used in paper
%%%%%%%%%%%%%%%%%%%%%%%%%%%%%%%%%%%%%%%%%%%%%%%%%%%%%%%%%%%

%%%%%%%%%%%%%%%%%%%%%%%%%%%%%%%%%%%%
% Utility commands
%%%%%%%%%%%%%%%%%%%%%%%%%%%%%%%%%%%%
\newcommand{\code}{\texttt}

\newcolumntype{Y}{>{\centering\arraybackslash}X}

%%%%%%%%%%%%%%%%%%%%%%%%%%%%%%%%%%%%
% supplemental material
%%%%%%%%%%%%%%%%%%%%%%%%%%%%%%%%%%%%

\newcommand{\supSecDataAvailability}{Section~2}
\newcommand{\supSecStreakMetric}{Section~5}
\newcommand{\supSecRepeatedSigAnalysis}{Section~7}
\newcommand{\supSecContextSigAnalysis}{Section~8}
\newcommand{\supSecBooleanCalcPrefixAnalysis}{Section~9}
\newcommand{\supSecChangingSigAnalysis}{Section~10}
\newcommand{\supSecBooleanCalcPostfixAnalysis}{Section~11}
\newcommand{\supSecHandcodedPrograms}{Section~12}

%%%%%%%%%%%%%%%%%%%%%%%%%%%%%%%%%%%%
% Evolution parameters
%%%%%%%%%%%%%%%%%%%%%%%%%%%%%%%%%%%%
\newcommand{\MutRateInstArgSub}{0.001}

\newcommand{\MutRateInstSub}{0.001}
\newcommand{\MutRateInstIns}{0.001}
\newcommand{\MutRateInstDel}{0.001}
\newcommand{\MutRateInstPoint}{0.001}

\newcommand{\MutRateSeqSlip}{0.05}

\newcommand{\MutRateFuncDup}{0.05}
\newcommand{\MutRateFuncDel}{0.05}

\newcommand{\MutRateInstTagBF}{0.0001}
\newcommand{\MutRateFuncTagBF}{0.0001}
\newcommand{\MutRateTagBF}{0.0001}

\newcommand{\MaxFuncCnt}{256}
\newcommand{\MaxFuncInstCnt}{128}

\newcommand{\MinArgValue}{\nobreakdash-4}
\newcommand{\MaxArgValue}{4}

%%%%%%%%%%%%%%%%%%%%%%%%%%%%%%%%%%%%
% Repeated-signal task parameters
%%%%%%%%%%%%%%%%%%%%%%%%%%%%%%%%%%%%
\newcommand{\RepeatSigTaskCpuTimePerSignal}{128}

%%%%%%%%%%%%%%%%%%%%%%%%%%%%%%%%%%%%
% Contextual-signal task parameters
%%%%%%%%%%%%%%%%%%%%%%%%%%%%%%%%%%%%
\newcommand{\ContextSigTaskCpuTimePerSignal}{128}

%%%%%%%%%%%%%%%%%%%%%%%%%%%%%%%%%%%%
% Boolean logic calculator task parameters
%%%%%%%%%%%%%%%%%%%%%%%%%%%%%%%%%%%%
\newcommand{\CalculatorTaskCpuTimePerSignal}{128}

%%%%%%%%%%%%%%%%%%%%%%%%%%%%%%%%%%%%
% Repeated-signal task results
%%%%%%%%%%%%%%%%%%%%%%%%%%%%%%%%%%%%
% FROM - 2020-11-25-rep-sig
%%% TWO SIGNAL %%%
\newcommand{\RSPTwoSigRegOnlySolutionCnt}{200}
\newcommand{\RSPTwoSigMemOnlySolutionCnt}{137}
\newcommand{\RSPTwoSigBothSolutionCnt}{200}
%%% FOUR SIGNAL %%%
\newcommand{\RSPFourSigRegOnlySolutionCnt}{200}
\newcommand{\RSPFourSigMemOnlySolutionCnt}{8}
\newcommand{\RSPFourSigBothSolutionCnt}{200}
%%% EIGHT SIGNAL %%%
\newcommand{\RSPEightSigRegOnlySolutionCnt}{198}
\newcommand{\RSPEightSigMemOnlySolutionCnt}{0}
\newcommand{\RSPEightSigBothSolutionCnt}{198}
%%% SIXTEEN SIGNAL %%%
\newcommand{\RSPSixteenSigRegOnlySolutionCnt}{73}
\newcommand{\RSPSixteenSigMemOnlySolutionCnt}{0}
\newcommand{\RSPSixteenSigBothSolutionCnt}{74}

% --- (KO) STRATEGIES (condition=regulation-augmented/memory+regulation) ---
%%% TWO SIGNAL %%%
\newcommand{\RSPTwoSigRegOnlyStrategyCnt}{188}
\newcommand{\RSPTwoSigMemOnlyStrategyCnt}{11}
\newcommand{\RSPTwoSigBothStrategyCnt}{1}

\newcommand{\RSPTwoSigRegRequiredStrategyCnt}{189}
\newcommand{\RSPTwoSigUnsolvedCnt}{0}

%%% FOUR SIGNAL %%%
\newcommand{\RSPFourSigRegOnlyStrategyCnt}{177}
\newcommand{\RSPFourSigMemOnlyStrategyCnt}{0}
\newcommand{\RSPFourSigBothStrategyCnt}{23}

\newcommand{\RSPFourSigRegRequiredStrategyCnt}{200}
\newcommand{\RSPFourSigUnsolvedCnt}{0}

%%% EIGHT SIGNAL %%%
\newcommand{\RSPEightSigRegOnlyStrategyCnt}{183}
\newcommand{\RSPEightSigMemOnlyStrategyCnt}{0}
\newcommand{\RSPEightSigBothStrategyCnt}{15}

\newcommand{\RSPEightSigRegRequiredStrategyCnt}{198}
\newcommand{\RSPEightSigUnsolvedCnt}{2}

%%% SIXTEEN SIGNAL %%%
\newcommand{\RSPSixteenSigRegOnlyStrategyCnt}{71}
\newcommand{\RSPSixteenSigMemOnlyStrategyCnt}{0}
\newcommand{\RSPSixteenSigBothStrategyCnt}{3}

\newcommand{\RSPSixteenSigRegRequiredStrategyCnt}{74}
\newcommand{\RSPSixteenSigUnsolvedCnt}{126}
%%%%%%%%%%%%%%%%%%%%%%%%%%%%%%%%%%%%
% Contextual-signal task results
%%%%%%%%%%%%%%%%%%%%%%%%%%%%%%%%%%%%
% From - 2020-11-27

\newcommand{\ContextFourSigMemOnlySolutionCnt}{173}
\newcommand{\ContextFourSigBothSolutionCnt}{200}

% --- (KO) STRATEGIES (condition=regulation-augmented/memory+regulation) ---
\newcommand{\ContextFourSigMemOnlyStrategy}{0}
\newcommand{\ContextFourSigRegOnlyStrategy}{105}
\newcommand{\ContextFourSigBothStrategy}{95}

%%%%%%%%%%%%%%%%%%%%%%%%%%%%%%%%%%%%
% Changing-signal task results
%%%%%%%%%%%%%%%%%%%%%%%%%%%%%%%%%%%%
% From - 

\newcommand{\ChangingSigSixteenSigMemOnlySolutionCnt}{X}
\newcommand{\ChangingSigSixteenSigBothSolutionCnt}{X}

%%%%%%%%%%%%%%%%%%%%%%%%%%%%%%%%%%%%
% Boolean Logic Calculator (PREFIX) results
%%%%%%%%%%%%%%%%%%%%%%%%%%%%%%%%%%%%
% From - 

\newcommand{\BoolCalcPrefixMemOnlySolutionCnt}{X}
\newcommand{\BoolCalcPrefixBothSolutionCnt}{X}

%%%%%%%%%%%%%%%%%%%%%%%%%%%%%%%%%%%%
% Boolean Logic Calculator (POSTFIX) results
%%%%%%%%%%%%%%%%%%%%%%%%%%%%%%%%%%%%
% From - 

\newcommand{\BoolCalcPostfixMemOnlySolutionCnt}{X}
\newcommand{\BoolCalcPostfixBothSolutionCnt}{X}

% Genetic programming and evolvable machines abstract guidelines:
% - 150 to 250 words
% -  The abstract should not contain any undefined abbreviations or unspecified references.

\begin{abstract}

We introduce and experimentally demonstrate the utility of tag-based genetic regulation, a new genetic programming (GP) technique that allows programs to dynamically adjust which code modules to express.
Tags are evolvable labels that provide a flexible mechanism for referencing code modules. 
Tag-based genetic regulation extends existing tag-based naming schemes to allow programs to ``promote'' and ``repress'' code modules in order to alter expression patterns.
This extension allows evolution to structure a program as a gene regulatory network where modules are regulated based on instruction executions.
We demonstrate the functionality of tag-based regulation on a range of program synthesis problems. 
We find that tag-based regulation improves problem-solving performance on context-dependent problems; that is, problems where programs must adjust how they respond to current inputs based on prior inputs.
Indeed, the system could not evolve solutions to some context-dependent problems until regulation was added.
Our implementation of tag-based genetic regulation is not universally beneficial, however.
We identify scenarios where the correct response to a particular input never changes, rendering tag-based regulation an unneeded functionality that can sometimes impede adaptive evolution.
Tag-based genetic regulation broadens our repertoire of techniques for evolving more dynamic genetic programs and can easily be incorporated into existing tag-enabled GP systems.

% We also observe that our implementation of tag-based genetic regulation can impede adaptive evolution when expected outputs are not context-dependent (\textit{i.e.}, the correct response to a particular input remains static over time). 

% Please provide 4 to 6 keywords which can be used for indexing purposes.

\end{abstract}

\section{Introduction}

% - Introduce genetic programming and automatic program synthesis + state what we do -
Genetic programming (GP) applies the natural principles of evolution to automatically synthesize programs rather than writing them by hand.
Indeed, the promise of automating computer programming has motivated advances in GP since its early successes in the 1980s \citep{cramer_representation_1985,forsyth_beagle_1981,koza_hierarchical_1989}. 
Just as human software developers have access to a dazzling array of programming languages, each specialized for solving different types of problems, GP features many ways to represent evolvable programs.
Each representation features different programmatic elements that vary in their syntax, organization, interpretation, and evolution.
These differences can dramatically influence the types of computer programs that can be evolved, and as such, influence a representation's problem-solving range \citep{hintze_buffet_2019,wilson_comparison_2008}. 
Here, we introduce and experimentally demonstrate tag-based module regulation for genetic programming, allowing us to more easily evolve programs capable of dynamically regulating responses to inputs over time.

%%%%%%%%%%%% lit to cite
% - TO READ (for different representations = different problem-solving):
%   - A Comparison of Cartesian Genetic Programming and Linear Genetic Programming
%%%%%%%%%%%%

% - Introduce value of software 'plasticity', link plasticity and modularity - 
Nearly all software applications are capable of conditionally responding to inputs.
For example, each input button on a calculator triggers a different software response; or, in the Small or Large problem from the Helmuth and Spector's automatic program synthesis benchmark suite \citep{helmuth_general_2015}, programs must output different classifications (``small'', ``large'', or ``neither'') depending on a numeric input value.
Just like such conditional logic is inherent in any non-trivial software, so to is it ubiquitous in biological organisms where it is referred to as ``plastic'' behavior or ``phenotypic plasticity.''
%Just like these dynamic responses are inherent in any non-trivial software, so to are they ubiquitous in natural organisms where they are referred to as ``plastic'' behaviors or ``phenotypic plasticity''.

% - link modular software design to improved software 'plasticity' -
Modular software design---that is, designs that promote the partitioning and reusability of functional units---is fundamental to good software development practices; this principle is all the more true in producing programs capable of complex ``plasticity.'' 
% Modular software design --- that is, designs that enable the separability and reusability of functional units --- is fundamental to good software development practices; this principle is all the more true in producing programs capable of complex ``plasticity''. 
By modularizing code (\textit{e.g.}, into functions, classes, libraries, \textit{etc.}), software developers can craft customized responses to inputs by composing relevant modules.  
These modules can each contain segments of code whose functionality would otherwise need to be reinvented for each response.
Likewise, modularity appears to be critical in natural genomes \citep{wagner_road_2007} as well as artificial evolving systems \citep{huizinga_does_2016}.
Moreover, evidence in these evolving systems suggests that modularity can improve the capacity for effective plasticity to arise \citep{ellefsen_neural_2015,londe_phenotypic_2015}. % and problem solving success?

Developing GP systems that facilitate the evolution of modular program architectures has long captured the attention of the genetic programming community. 
Koza introduced Automatically Defined Functions (ADFs) where callable functions can evolve as separate branches of GP syntax trees \citep{koza_genetic_1992,koza_genetic_1994}.
Angeline and Pollack developed compression and expansion genetic operators to automatically modularize existing code into libraries of parameterized subroutines \citep{angeline_evolutionary_1992}. 
Since these foundational advances, significant efforts have been made to allow GP representations to incorporate internal modules 
(\textit{e.g.},
\citep{spector_adm_1996,oneill_grammar_2000,binard_abstraction_2007,walker_automatic_2008,spector_tag-based_2011,spector_tag-based_2012,lalejini_evolving_2018}),
to measure (and select for) modularity in evolving programs 
(\textit{e.g.}, \citep{krawiec_functional_2009,saini_modularity_2019,saini_using_modularity_2020}),
and to build ``libraries'' of reusable code modules accessible to evolving populations of programs
(\textit{e.g.}, \citep{banscherus_hierarchical_2001,keijzer_run_2004,keijzer_undirected_2005,rosca_learning_1994}).

% - Introduce more complicated form of plasticity -
These innovations have improved the ability of GP systems to link modules together to solve problems, thus improving their prospects as general-purpose tools for automatic program synthesis.
In existing GP work, links between modules, however, are typically hard coded and static during program execution. 
Less is known for how to evolve programs that can adjust module associations on the fly.
For many types of problems, the appropriate set of modules to execute in response to a particular input changes over time.
This requires programs to continuously adjust associations between inputs and modular responses based on context.
For example, the computations that occur on a calculator after pressing the ``equals'' button are \textit{context-dependent}; that is, they depend on the set of operators and operands (\textit{i.e.}, inputs) previously provided. 
To achieve this design pattern, programs must internally track contextual information and typically regulate responses using explicit flow control directives (such as if-statements).
Our goal is to evolve programs that dynamically regulate modules during execution to more effectively solve context-dependent problems.  
To reach this goal, we draw inspiration from gene regulatory networks (both natural and artificial) to augment how program modules are called in GP.

% - Introduce what we did, summarize our findings -
Here, we propose to facilitate dynamic module composition by introducing tag-based module regulation for genetic programming. 
We extend existing tag-based naming schemes to allow programs to dynamically adjust associations between references and code modules. 
We experimentally demonstrate our implementation of tag-based genetic regulation in the context of SignalGP \citep{lalejini_evolving_2018}; however, our approach is immediately applicable to any existing tag-enabled GP system, such as tag-addressed Run Transferable Libraries \citep{keijzer_run_2004} or PushGP \citep{spector_tag-based_2011}.
We add ``regulation'' instructions to SignalGP that can adjust (\textit{i.e.}, promote or repress) which code modules respond to input signals and internal calls.
This extension allows evolution to structure a program as a gene regulatory network where genes are program modules and program instructions mediate regulation.
We show that module regulation improves problem-solving performance on problems where responses to particular inputs change depending on prior context (\textit{e.g.}, prior inputs). 
We also observe that our implementation of tag-based regulation can sometimes impede adaptive evolution when outputs are not context-dependent. 
% (\textit{i.e.}, the correct response to a particular input does not change over time).

\section{Specifying Modules with Tag-based Referencing}
\label{sec:tag-based-referencing}
% Alternatively, just 'Tag-based Referencing'

% - motivation for evolvable names -
All programming representations that support modularizing code into functions or libraries define mechanisms for labeling and subsequently referencing modules. 
In traditional software development, programmers hand label modules and reference a particular module using its assigned label.
Programmers must precisely name the module they intend to reference; imprecision typically results in incorrect outputs or a syntax error. 
This mechanism for referencing modules allows for an arbitrarily large space of possible module names and is intentionally brittle, ensuring programs are either interpreted by a computer exactly as written or not interpreted at all. 
Requiring genetic programming systems to adhere to these traditional approaches to module referencing is not ideal. 
Mutation operators must either ensure that mutated labels are syntactically valid, or else cope with an abundance of broken code.
These choices result in either a search space that is overly constrained or one that is rugged and difficult to navigate \citep{rasmussen_coreworld_1990}.

% - tag-based referencing (definition) -
Inspired by Holland's use of ``tags'' to facilitate binding and aggregation in complex adaptive systems \citep{holland_concerning_1990,holland_effect_1993}, %XXX implemented [citations] and then 
Spector \textit{et al.} generalized the use of tags to label and refer to program modules in GP \citep{spector_whats_2011,spector_tag-based_2011}.
Tags are evolvable labels that can be mutated, and the similarity (or dissimilarity) between any two tags can be quantified. 
Tags are most commonly represented as floating point or integer numeric values \citep{keijzer_run_2004,spector_tag-based_2011} or as bit strings \citep{lalejini_evolving_2018}.
Like traditional naming schemes, tags can provide an arbitrarily large address space.
Unlike traditional naming schemes, however, tags allow for \textit{inexact} addressing. 
A referring tag targets the tagged entity (\textit{e.g.}, a module) with the \textit{closest matching} tag; 
this ensures that all possible tags are valid references.
Further, mutations to tags do not necessarily invalidate existing references.
For example, mutating a referring tag will have no phenotypic effect if those mutations do not change which target tag is matched. 
As such, mutating tag-based names is not necessarily catastrophic to program functionality, allowing the labeling and use of modularized code fragments to incrementally co-evolve~\citep{spector_tag-based_2011}.

% - history/background for tag-based referencing -
Tag-based referencing has long been used to expand the capabilities of genetic programming systems.
Keijzer \textit{et al.} created run transferable libraries of tag-addressable functions using successful code segments evolved in previous GP runs \citep{keijzer_run_2004,keijzer_undirected_2005}.
Evolving programs (represented as program trees) contained dynamically-linked nodes that used tag-based referencing to call library functions.
These tag-addressed libraries were updated between runs and did not co-evolve with programs.

Spector \textit{et al.} augmented PushGP with tag-based referencing, allowing tag-addressable code modules to evolve \textit{within} a program \citep{spector_tag-based_2011}.
Spector \textit{et al.} found that tags provided a flexible mechanism for modularization that allowed tag-enabled programs to better scale with problem size. 
Additionally, Spector \textit{et al.} expanded tag-based modules beyond PushGP, successfully applying the technique to tree-based GP \citep{spector_tag-based_2012}.

Lalejini and Ofria further extended tag-based naming to linear GP.  
Their SignalGP system broadens the application of tags to facilitate the evolution of event-driven programs \citep{lalejini_evolving_2018,lalejini_what_2019}. 
In SignalGP, tagged modules are called internally or triggered in response to tagged events (\textit{e.g.}, events generated by other agents or the environment).
More recently, Lalejini and Ofria demonstrated the use of tags to label memory positions in GP, enabling programs to define and use evolvable variable names \citep{lalejini_tag-accessed_2019}.
This tag-based memory implementation did not substantively affect problem-solving performance; however, tag-based addressing features a larger addressable memory space than more traditional register-based memory approaches in GP.
% @AML: Anything from DISHTINY we should cite?

\section{Tag-based Genetic Regulation}
\label{sec:tag-based-genetic-regulation}

Here, we allow programs to use tag-based referencing to dynamically regulate module execution.
To achieve this, we draw inspiration from both natural and artificial gene regulatory networks. 
We demonstrate that this approach promotes more effective solutions for context-dependent problems.

% - gene regulation background, briefly -
Gene regulatory networks represent the complex interactions among genes, transcription factors, and signals from the environment that, together, control gene expression \citep{banzhaf_artificial_2015}.
Gene regulation allows for feedback loops so that prior events can continue to influence future expression in flexible and nuanced ways. 
Gene regulation underlies most important biological processes, including cell differentiation, metabolism, the cell cycle, and signal transduction \citep{karlebach_modelling_2008}.
The role of gene regulatory networks in sustaining complex life has inspired varied and abundant computational models of these networks \citep{cussat-blanc_artificial_2019,karlebach_modelling_2008}.

% ---- BEGIN REVISION ----
Artificial gene regulatory networks have been used to study how natural gene regulation evolves \citep{aldana_robustness_2007,crombach_evolution_2008,draghi_evolutionary_2009} and as a tool in evolutionary computation to solve challenging control problems (as reviewed by \citep{cussat-blanc_artificial_2019}).
Evolved artificial gene regulatory networks have even been used as indirect encoders, providing a developmental phase to translate genomes into programs \citep{banzhaf_artificial_2003,lopes_regulatory_2012} or neural networks \citep{kowaliw_using_2014}.
La Cava \textit{et al.} demonstrated a form of \textit{epigenetic} regulation for genetic programming where ``gene'' activation and silencing is learned each generation \citep{la_cava_genetic_2015,la_cava_inheritable_2015}; however, the programs themselves did not have direct control over these regulatory elements.
% @AML: new sentence below
Inspired by chromatin remodeling in biological cells, Turner \textit{et al.} introduced artificial epigenetic networks that allow for the regulation (\textit{i.e.}, the addition or removal) of internal network components  \citep{turner_artificial_2017}; such topological self-modification improved problem-solving success for dynamical control problems.
% ---- END REVISION ----

% ----- REVISIONS BEGIN -----
% - Issues
%  - Hard to understand (the next paragraph). Not sufficient information about how SignalGP executes its linear programs.
%  - Use terms like 'down-regulated', 'up-regulated', and 'tag-match score' without defining what they mean.
%  - Also, I can't figure out what the sentence "Promoter and repressor instructions use their associated tags to target modules in the program; these instructions use unregulated tag-based referencing to identify modules to avoid unrecoverable states where, for example, a down-regulated function could not be subsequently up-regulated."
%  - Since this paragraph details the main contributions of this paper, please edit or rewrite it to be significantly clearer to those not familiar with SignalGP.
% - Notes
%   - I might have gone too far in some places, making things too verbose/repetitive. 

We aim to incorporate gene regulatory network-inspired methodology to allow programs to dynamically adjust which module is triggered by a particular call based on not just current inputs, but also prior inputs.
We achieved this goal by instantiating gene regulatory networks using tag-based referencing.
Specifically, we implemented tag-based genetic regulation in the context of the linear GP system SignalGP \citep{lalejini_evolving_2018}, which is described in further detail in Section \ref{sec:methods:signalgp}.
Here, we describe tag-based genetic regulation in terms of our SignalGP-based implementation; however, our overall approach is immediately applicable to each of the tag-enabled systems described in Section \ref{sec:tag-based-referencing} and can be easily incorporated into any genetic programming representation.

Briefly, programs in SignalGP are composed of tag-addressed modules (\textit{i.e.}, functions), each of which contain a linear sequence of instructions. 
Each instruction has arguments, including an evolvable tag that can be used to identify and call a tag-addressed module.
When a referring tag (\textit{e.g.}, from an instruction) is used to look up a tag-addressed module, all modules in that program are ranked according to a tag-matching score. 
A tag-matching score quantifies the quality of the reference between a referring tag and a module's tag; we always select the module with best reference quality (\textit{i.e.}, the highest tag-match score with the referring tag).
When a module is called, it is executed procedurally, instruction-by-instruction, in the same way as in a conventional linear GP system. 

\begin{figure}[!ht]
    \centering
    \includegraphics[width=\textwidth]{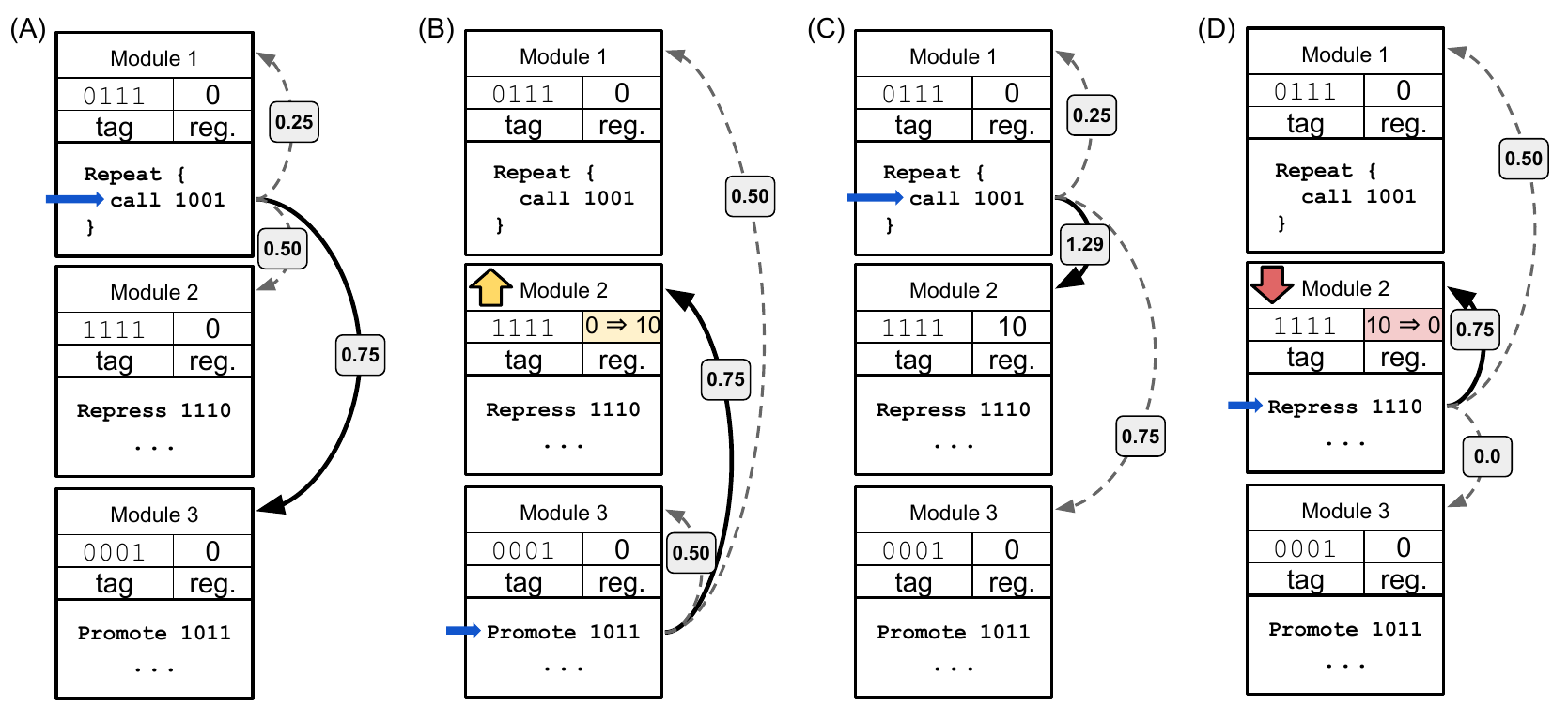}
    \caption{\small
    \textbf{Tag-based genetic regulation example.}
    This example depicts a simple oscillating regulatory network instantiated using tag-based regulation.
    In this example, tags are length-4 bit strings. 
    The ``raw'' match score between two tags equals the number of matching bits between them.
    Regulation (reg.) modifies match scores for ``\code{call}'' instructions according to Equation \ref{equ:reg-transform}.
    First (A), the \code{call 1001} in Module 1 executes, triggering Module 3. 
    Next (B), Module 3 is executed, promoting Module 2. 
    After Module 3 returns, the \code{call 1001} in Module 1 executes again (C); however, Module 2's promotion causes it to be triggered instead of Module 3. 
    Finally (D), Module 2 executes and represses itself, resetting its regulatory modifier to 0. 
    }
    \label{fig:regulation-example}
\end{figure}

% Revision
We modified SignalGP in two ways to implement tag-based genetic regulation: 

\begin{enumerate}
    \item We added a ``regulatory modifier'' value (represented as a floating point value) to all tag-addressed modules. A module's regulatory modifier adjusts how well that module will match to referring tags, and thus, modifies the likelihood it will be referenced.
    \item We supplemented the instruction set with promoter and repressor instructions that, when executed, adjust a target module's regulatory modifier. 
\end{enumerate} 

When a program begins execution, each internal module initially has no regulatory modification.\footnote{
    Alternatively, allowing programs to inherit their parent's regulatory modifiers can provide a simple model of epigenetics.
}
When a promoter or repressor instruction is executed, its associated tag identifies which module should be regulated using tag-based referencing.
Promoter instructions increase a target module's regulatory modifier, which increases the module's tag-match score with subsequent references (according to equation \ref{equ:reg-transform} below) and thus increases the module's chances of being referenced.
Repressor instructions have the opposite effect.
Regulatory modifiers can be configured to persist over a program's entire execution or passively decay over time.
% passively decay over time or persist over a program's entire execution.

When determining which module to call at runtime, each module's tag-match score is a function of how well the module's tag matches the call instruction's tag as modified by the module's regulatory value.
% @AML: This next sentence (or two) is in rough shape, need streamlining.
If a module's regulatory modifier has been sufficiently decreased by repressor instructions, it is possible that the module will no longer be able to be referenced, as its regulated tag-match score will always be lower than at least one other program module.
We must ensure that this situation does not create an unrecoverable regulatory state and that such a fully repressed module can always be restored.
As such, promoter and repressor instructions use \textit{unregulated} tag-based referencing to identify which modules they regulate; that is, we do not apply regulatory modifiers to tag-based references made by promoter and repressor instructions. 
This ensures that no matter how much a particular module has been repressed, subsequent promoter instructions can increase its regulatory modifier.
Figure \ref{fig:regulation-example} gives a simplified example of how promoter and repressor instructions can dynamically adjust module execution over time.

We have implemented a toolbox of interchangeable methods for applying regulation to tag-matching scores in the Empirical library \citep{charles_ofria_2020_empirical}. 
Here, we use a simple exponential function to apply a module's regulation modifier to its tag-match score calculations: 

\begin{equation}
M_{r}(t_q, t_m, R_m) = M(t_q, t_m) \times b^{R_m}
\label{equ:reg-transform}
\end{equation}

\noindent
$R_m$ specifies the module's regulation modifier, which is under the direct control of the evolving programs.
$M_r$ is the regulation-adjusted match score between a querying tag ($t_q$) and the module's tag ($t_m$).
$M$ is a function that gives the baseline, unadjusted match score between the querying tag and module tag.
If tags are represented as floating point values, $M$ can be as simple as the absolute difference between the two tags.
%; in this work, tags are represented as bit strings, and $M$ is computed using the Streak metric \citep{downing_intelligence_2015}.
The strength of regulation is determined by the constant, $b$ (set to $1.1$ in this work). 

\begin{figure}[htbp]
    \centering
    \includegraphics[width=0.6\columnwidth]{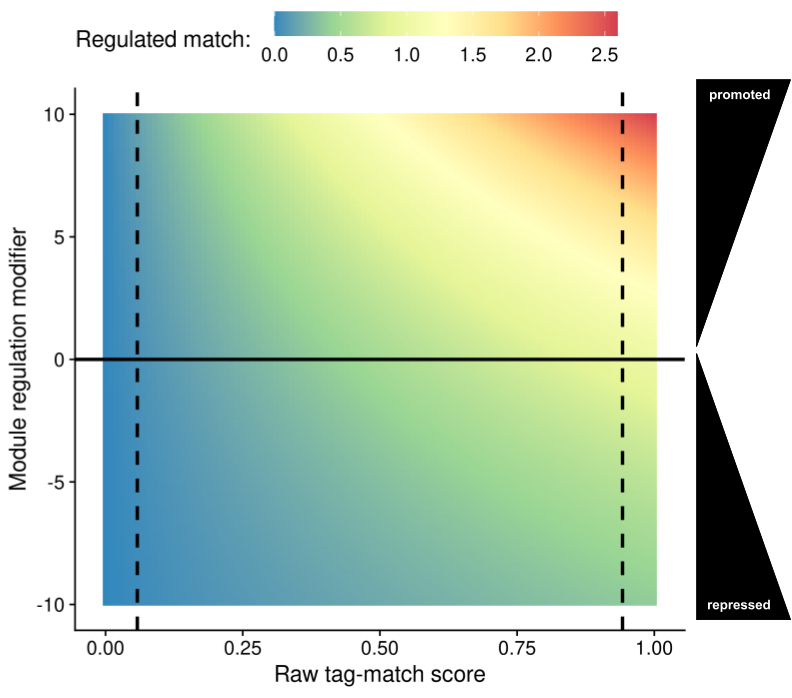}
    \caption{\small
    \textbf{Regulated tag-match score as a function of raw tag-match score and regulatory modifier values according to Equation \ref{equ:reg-transform}.}
    The horizontal black line indicates a neutral regulatory state; 
    repressed states are below the line, and promoted states are above the line.
    We expect the raw tag-match score (calculated using the Streak similarity metric, which is described later in Section \ref{sec:methods:signalgp}) of 90\% of random pairs of tags to fall between the two dashed vertical lines; to compute the location of these lines, we generated $10^{5}$ pairs of random tags and found the region that contained the middle 90\% of raw tag-matching scores.
    }
    \label{fig:exponential-regulation-function}
\end{figure}

When determining which module to reference, each candidate module's $M_r$ is computed, and the module with the highest $M_r$ value is chosen. 
Intuitively, modules with $R_m < 0$ are down-regulated (\textit{i.e.}, in a repressed state), modules with $R_m > 0$ are up-regulated (\textit{i.e.}, in a promoted state), and modules with $R_m = 0$ are unmodified by regulation. 
That is, down-regulated modules have lower tag-match scores than they otherwise would without regulation, and up-regulated modules have higher tag-match scores than they otherwise would without regulation.
Figure \ref{fig:exponential-regulation-function} gives a visual representation of Equation \ref{equ:reg-transform}.

In preliminary experiments, we tested several different methods of implementing regulation (including additive, multiplicative, and the current exponential techniques).
We found no evidence for any one method performing substantially better than the others.
Future work will more thoroughly explore the potential effects of different regulation mechanisms. 

% ----- REVISIONS END -----

\section{Methods}
\label{sec:methods}

We evaluated how tag-based genetic regulation faculties contribute to, and potentially detract from, the functionality of evolved genetic programs in the context of SignalGP.
First, we assessed the evolvability of our implementation of tag-based genetic regulation: 
can we evolve programs that rely on regulation to dynamically adjust their response to environmental conditions over time? 
Additionally, can tag-based genetic regulation improve problem-solving success on context-dependent problems?
We addressed these questions using the signal-counting and contextual-signal problems, diagnostic tasks that require context-dependent responses to an input signal.

Next, we assessed tag-based genetic regulation on the Boolean-logic calculator problem, a more challenging program synthesis problem that requires programs to perform Boolean logic computations in response to a sequence of input events that represent button presses on a simple calculator. 

% ---- BEGIN REVISION ----
Finally, we used the independent-signal problem to investigate the potential for genetic regulation to impede adaptive evolution by producing maladaptive plasticity. 
The independent-signal problem is a diagnostic that requires programs to associate distinct responses with each type of input; as such, programs do not need to change their response to particular input signals based on prior context.
Additionally, fitness evaluation in the independent-signal problem is imperfect: programs receive input signals in a random order, providing ample opportunity for erroneous regulation to impede adaptive evolution.

% fitness evaluation in the independent-signal problem is imperfect, providing ample opportunity for erroneous regulation to impede adaptive evolution.  

% The changing-signal task does not require programs to contextually change their response to particular inputs, providing ample opportunity for erroneous regulation to impede adaptive evolution.  

% ----- END REVISION ----

\subsection{SignalGP}
\label{sec:methods:signalgp}

% - High-level overview of SignalGP -
%   - event-driven computing
Here, we provide a general overview of SignalGP; see \citep{lalejini_evolving_2018} for a more in-depth description.
SignalGP defines a scheme for organizing and interpreting genetic programs to afford computational evolution access to the event-driven programming paradigm \citep{cassandras_event-driven_2014}.
In event-driven programs, software execution focuses on processing events (often in the form of messages from other processes, sensor alerts, or user actions).
In SignalGP, events (signals) trigger the execution of program modules (functions), facilitating efficient reactions to exogeneously- or endogeneously-generated signals.
For this work, program modules are represented as sequences of instructions; however, the SignalGP framework generalizes across a variety of program representations \citep{lalejini_what_2019}.

% - Slightly less high-level overview of SignalGP -
%   - organization of programs
%   - tags, tag-based referencing (in signalgp context)
%   - how signals trigger functions
Programs in SignalGP are explicitly modular, comprising a set of functions, each associating a tag with an instruction sequence.
SignalGP makes explicit the concept of events or \textit{signals}.
All signals contain a tag and any associated signal-specific data (\textit{e.g.}, numeric input values).
Because both signals and program functions are tagged, SignalGP determines the most appropriate function to process a signal using tag-based referencing: signals trigger the function with the closest matching tag.

% ------ BEGIN REVISIONS --------
% Comment: Please do two things to add clarity here: 
%  - (1) actually show the equation for the Streak metric, 
%  - (2) clearly define what you mean by "longest contiguously-matching" and "longest contiguously-mismatching". 
%  - For example, are these aligned by index, or are slight misalignments taken into account, for example like Levenshtein string distances? Also, what if two bit strings are identical -- is the mismatch length 0, and then what happens when you divide by 0?

In this work, we represent tags as 256-bit strings, and we quantify the similarity between any two tags using the Streak metric.
The Streak metric was originally proposed by Downing \citep{downing_intelligence_2015} and measures similarity between two bit strings in terms of the relationship between the lengths of the longest contiguously-matching and longest contiguously-mismatching substrings.\footnote{
We make a slight modification to Downing's matching procedure due to an error in its mathematical derivation, as detailed in the supplement \citep{lalejini_supplement}.
}
Specifically, we XOR the two bit strings and count the longest substring of all 0's in the first case or of all 1's in the second.
The equation below overviews how the Streak metric computes the similarity ($S$) between two tags ($t_q$ and $t_m$):

% \begin{equation*}
%     \mathrm{match score} = \frac{ p(\mathrm{match}) }{ p(\mathrm{mismatch}) + p(\mathrm{match}) } 
% \end{equation*}

\begin{equation*}
    S(t_q,t_m) = \frac{ p_{\mathrm{mismatch}}(t_q,t_m) }{ p_{\mathrm{mismatch}}(t_q,t_m) + p_{\mathrm{match}}(t_q,t_m) } 
\end{equation*}

\noindent
where $p_{\mathrm{match}}$ returns the probability of observing the measured length of the longest contiguously-matching substring between $t_q$ and $t_m$ by chance, and $p_{\mathrm{mismatch}}$ returns the probability of observing the measured length of the longest contiguously-mismatching substring between $t_q$ and $t_m$ by chance. 
Both our implementation and the mathematical equations for computing the Streak similarity between two bit strings can be found in supplemental material \supSecStreakMetric\ \citep{lalejini_supplement}.

% ------ END REVISIONS --------

% ----- BEGIN REVISIONS -----
% -- Signal-driven execution --
When a signal triggers a function, the function executes with the signal's associated data as input.
SignalGP programs can handle many signals simultaneously by processing and responding to each in parallel threads of execution.
Threads each contain local memory registers for performing computations.
Additionally, concurrently executing threads may interact by writing to and reading from a shared global memory buffer.
% @AML: new lines next
For this work, we guaranteed deterministic thread execution using a round robin scheduler to step each thread forward one step (\textit{i.e.}, one instruction) synchronously.

% ----- END REVISIONS -----

% - Instructions -
%   - see supplemental material for full instruction set!
The SignalGP instruction set allows programs to generate internal signals, broadcast external signals, and otherwise work in a tag-based context. 
In this work, each instruction contains one tag and three integer arguments. 
Arguments may modify the effect of an instruction, often specifying memory locations or fixed values.
For example, instructions may refer to and call internal program modules using tag-based referencing; when an instruction generates a signal (\textit{e.g.}, to be used internally or broadcast), the instruction's tag is used as the signal's tag.

%%%%%%%%%%%%%%%%%%%%%%%%%%%%%%%%%%%%%%%%%%%%%%%%%%%%%%%%%%%%%%%%%%%%%%%%%%%%%%%%%%%%%%%%%%%%%%
% Variables to make it easy to change table contents
%%%%%%%%%%%%%%%%%%%%%%%%%%%%%%%%%%%%%%%%%%%%%%%%%%%%%%%%%%%%%%%%%%%%%%%%%%%%%%%%%%%%%%%%%%%%%%

%%%%%%%%%% SetRegulator %%%%%%%%%%
\newcommand{\SetRegulatorName}{\code{SetRegulator+}}
\newcommand{\SetRegulatorArgs}{1}
\newcommand{\SetRegulatorTag}{Yes}
\newcommand{\SetRegulatorDescription}{Set the regulatory modifier of a target module to the value stored in an argument-specified memory register.}

%%%%%%%%%% SetRegulator- %%%%%%%%%%
\newcommand{\SetRegulatorMinusName}{\code{SetRegulator-}}
\newcommand{\SetRegulatorMinusArgs}{1}
\newcommand{\SetRegulatorMinusTag}{Yes}
\newcommand{\SetRegulatorMinusDescription}{Set the regulatory modifier of a target module to the negation of the value stored in an argument-specified memory register.}

\newcommand{\SetOwnRegulatorName}{\code{SetOwnRegulator+}}
\newcommand{\SetOwnRegulatorArgs}{1}
\newcommand{\SetOwnRegulatorTag}{No}
\newcommand{\SetOwnRegulatorDescription}{Set the regulatory modifier of the currently executing module to the value stored in an argument-specified memory register.}

\newcommand{\SetOwnRegulatorMinusName}{\code{SetOwnRegulator-}}
\newcommand{\SetOwnRegulatorMinusArgs}{1}
\newcommand{\SetOwnRegulatorMinusTag}{No}
\newcommand{\SetOwnRegulatorMinusDescription}{Set the regulatory modifier of the currently executing module to the negation of the value stored in an argument-specified memory register.}

\newcommand{\AdjRegulatorName}{\code{AdjRegulator+}}
\newcommand{\AdjRegulatorArgs}{1}
\newcommand{\AdjRegulatorTag}{Yes}
\newcommand{\AdjRegulatorDescription}{Add the value stored in an argument-specified memory register to the regulatory modifier of a target module.}

\newcommand{\AdjRegulatorMinusName}{\code{AdjRegulator-}}
\newcommand{\AdjRegulatorMinusArgs}{1}
\newcommand{\AdjRegulatorMinusTag}{Yes}
\newcommand{\AdjRegulatorMinusDescription}{Subtract the value stored in an argument-specified memory register to the regulatory modifier of a target module.}

\newcommand{\AdjOwnRegulatorName}{\code{AdjOwnRegulator+}}
\newcommand{\AdjOwnRegulatorArgs}{1}
\newcommand{\AdjOwnRegulatorTag}{No}
\newcommand{\AdjOwnRegulatorDescription}{Add the value stored in an argument-specified memory register to the regulatory modifier of the currently executing module.}

\newcommand{\AdjOwnRegulatorMinusName}{\code{AdjOwnRegulator-}}
\newcommand{\AdjOwnRegulatorMinusArgs}{1}
\newcommand{\AdjOwnRegulatorMinusTag}{No}
\newcommand{\AdjOwnRegulatorMinusDescription}{Subtract the value stored in an argument-specified memory-register to the regulatory modifier of the currently executing module.}

\newcommand{\ClearRegulatorName}{\code{ClearRegulator}}
\newcommand{\ClearRegulatorArgs}{0}
\newcommand{\ClearRegulatorTag}{Yes}
\newcommand{\ClearRegulatorDescription}{Reset the regulatory modifier of a target module.}

\newcommand{\ClearOwnRegulatorName}{\code{ClearOwnRegulator}}
\newcommand{\ClearOwnRegulatorArgs}{0}
\newcommand{\ClearOwnRegulatorTag}{No}
\newcommand{\ClearOwnRegulatorDescription}{Reset the regulatory modifier of the currently executing module.}

\newcommand{\SenseRegulatorName}{\code{SenseRegulator}}
\newcommand{\SenseRegulatorArgs}{1}
\newcommand{\SenseRegulatorTag}{Yes}
\newcommand{\SenseRegulatorDescription}{Load the value of a target module's regulatory modifier into an argument-specified memory register.}

\newcommand{\SenseOwnRegulatorName}{\code{SenseOwnRegulator}}
\newcommand{\SenseOwnRegulatorArgs}{1}
\newcommand{\SenseOwnRegulatorTag}{No}
\newcommand{\SenseOwnRegulatorDescription}{Load the value of the currently executing module's regulatory modifier into an argument-specified memory register.}

\newcommand{\IncRegulatorName}{\code{IncRegulator}}
\newcommand{\IncRegulatorArgs}{0}
\newcommand{\IncRegulatorTag}{Yes}
\newcommand{\IncRegulatorDescription}{Add one to the regulatory modifier of a target module. }

\newcommand{\IncOwnRegulatorName}{\code{IncOwnRegulator}}
\newcommand{\IncOwnRegulatorArgs}{0}
\newcommand{\IncOwnRegulatorTag}{No}
\newcommand{\IncOwnRegulatorDescription}{Add one to the regulatory modifier of the currently executing module.}

\newcommand{\DecRegulatorName}{\code{DecRegulator}}
\newcommand{\DecRegulatorArgs}{0}
\newcommand{\DecRegulatorTag}{Yes}
\newcommand{\DecRegulatorDescription}{Subtract one from the regulatory modifier of a target module.}

\newcommand{\DecOwnRegulatorName}{\code{DecOwnRegulator}}
\newcommand{\DecOwnRegulatorArgs}{0}
\newcommand{\DecOwnRegulatorTag}{No}
\newcommand{\DecOwnRegulatorDescription}{Subtract one from the regulatory modifier of the currently executing module.}

%%%%%%%%%%%%%%%%%%%%%%%%%%%%%%%%%%%%%%%%%%%%%%%%%%%%%%%%%%%%%%%%%%%%%%%%%%%%%%%%%%%%%%%%%%%%%%
% Table
%%%%%%%%%%%%%%%%%%%%%%%%%%%%%%%%%%%%%%%%%%%%%%%%%%%%%%%%%%%%%%%%%%%%%%%%%%%%%%%%%%%%%%%%%%%%%%
% \setlength{\tabcolsep}{16pt}
\renewcommand{\arraystretch}{1.5}
\begin{table}[htbp]
    \small
    \centering
    \rowcolors{2}{gray!25}{white}
    \begin{tabular}{l  p{0.6\columnwidth}}
        \rowcolor{gray!50}
        % \toprule
        \hline
        \textbf{Instruction} & 
        \textbf{Description} 
        \\ \hline 
        
        \SetRegulatorName & 
        \SetRegulatorDescription 
        \\ %\hline
        
        \SetRegulatorMinusName & 
        \SetRegulatorMinusDescription 
        \\ %\hline
        
        \SetOwnRegulatorName &
        \SetOwnRegulatorDescription 
        \\ %\hline
        
        \SetOwnRegulatorMinusName & 
        \SetOwnRegulatorMinusDescription 
        \\ %\hline
        
        \AdjRegulatorName & 
        \AdjRegulatorDescription 
        \\ %\hline
        
        \AdjRegulatorMinusName & 
        \AdjRegulatorMinusDescription 
        \\ %\hline
        
        \AdjOwnRegulatorName & 
        \AdjOwnRegulatorDescription 
        \\ %\hline
        
        \AdjOwnRegulatorMinusName & 
        \AdjOwnRegulatorMinusDescription 
        \\ %\hline
        
        \ClearRegulatorName & 
        \ClearRegulatorDescription 
        \\ %\hline
        
        \ClearOwnRegulatorName & 
        \ClearOwnRegulatorDescription
        \\ %\hline
        
        \SenseRegulatorName & 
        \SenseRegulatorDescription 
        \\ %\hline
        
        \SenseOwnRegulatorName & 
        \SenseOwnRegulatorDescription
        \\ %\hline
        
        \IncRegulatorName & 
        \IncRegulatorDescription 
        \\ %\hline
        
        \IncOwnRegulatorName & 
        \IncOwnRegulatorDescription 
        \\ %\hline
        
        \DecRegulatorName & 
        \DecRegulatorDescription 
        \\ %\hline
        
        \DecOwnRegulatorName & 
        \DecOwnRegulatorDescription
        \\
        \hline
        % \bottomrule
    \end{tabular}
    \caption{\small 
    \textbf{Regulatory instructions used in this work.} 
    We include (+) and (-) instruction variants to ensure that positive and negative regulation values are equally probable.
    }
    \label{tab:regulation-instructions}
\end{table}

Previous work has demonstrated that SignalGP facilitates the evolution of event-driven programs capable of identifying and responding to many \textit{distinct} signals \citep{lalejini_what_2019}.
% However, without access to regulation, SignalGP requires programs to use procedural mechanisms (\textit{e.g.}, if statements) to adjust how they respond to a particular signal over time.
However, without access to regulation, SignalGP requires programs to track context in memory and use procedural mechanisms (\textit{e.g.}, if statements) to adjust how they respond to a particular signal over time based on stored context.
Here, we apply tag-based genetic regulation to SignalGP (as described in Section \ref{sec:tag-based-genetic-regulation}).
We supplemented the instruction set with regulatory instructions (Table \ref{tab:regulation-instructions}) that use tag-based referencing to target internal functions. 
In this work, we apply regulation to function references using Equation \ref{equ:reg-transform}. 
Our full instruction set, including descriptions of each instruction, can be found in our supplemental material \citep{lalejini_supplement}.

\subsubsection{Evolution}

In this work, we propagated programs asexually, and we applied mutations to offspring.
The parent-selection method varied across experiments.
Programs were variable-length: each program contained up to \MaxFuncCnt\ modules, and each module contained up to \MaxFuncInstCnt\ instructions.

% ---- BEGIN REVISION -----

We applied single-instruction substitution, insertion, and deletion mutations each at a per-instruction rate of \MutRateInstPoint. 
Additionally, we applied a `slip' mutation operator \citep{lalejini_gene_2017} that could duplicate or delete entire sequences of instructions at a per-module rate of \MutRateSeqSlip. 
We mutated numeric instruction arguments at a per-argument rate of \MutRateInstArgSub, and we limited numeric arguments to values between \MinArgValue\ and \MaxArgValue.
When a numeric argument mutated, we randomized the argument's value to a valid integer between~\MinArgValue\ and~\MaxArgValue.
We mutated instruction- and module-tags at a per-bit rate of \MutRateTagBF.
We applied whole-module duplication and deletion operators at a per-module rate of \MutRateFuncDup, allowing the number of modules in a program to evolve.

% ---- END REVISION -----

% \subsection{Repeated-signal Problem}
% \label{sec:methods:repeated-signal-problem}
\subsection{Signal-counting Problem}
\label{sec:methods:signal-counting-problem}

% ----- BEGIN REVISION -----
% - Issue: does a program simply have to execute the corresponding Response-i instruction, or do the Response-i instructions output some other value? I think the former, but this could be clearer.
The signal-counting problem requires programs to continually change their response to an environmental signal, producing the appropriate output each of the $K$ times that signal is repeated.
Programs output responses by executing one of $K$ response instructions. 
For example, if a program receives two signals from the environment during evaluation (\textit{i.e.}, $K=2$), the program should execute \code{Response-1} after the first signal and \code{Response-2} after the second signal; aside from executing the correct response instruction, no other output is necessary after receiving an environmental signal.

% ----- END REVISION -----

% ----- BEGIN REVISION -----
% - Issue: what is a time step?
We provide programs \RepeatSigTaskCpuTimePerSignal\ time steps to respond to each environmental signal.
% @AML: new sentence next
During each time step, each of a program's active threads execute a single instruction.
Once the allotted time expires or the program outputs a response, the program's threads of execution reset, resulting in a loss of all thread-local memory; \textit{only} the contents of the global memory buffer and each program module's regulatory state persist.
The environment then produces the next signal (identical to each previous environmental signal) to which the program may respond.
A program \textit{must} use the global memory buffer or genetic regulation to correctly shift its response to each subsequent environmental signal. 
Evaluation continues in this way until the program correctly responds to each of the $K$ environmental signals or until the program executes an incorrect response.
A program's fitness equals the number of consecutive correct responses given during evaluation, and a program is considered a solution if it correctly responds to all $K$ environmental signals.

% ----- END REVISION ----

\subsubsection{Experimental Design}

The signal-counting problem is explicitly designed to 
(1) evaluate if tag-based genetic regulation can be evolved to dynamically adjust which modules execute in response to a \textit{repeated} input type 
and (2) assess the problem-solving success of a regulation-enabled GP system relative to an otherwise identical GP system with regulation disabled. 
We compared programs evolved in a regulation-on treatment to those evolved in a regulation-off control.
In the control treatment, we used an identical instruction set where regulation instructions were altered to behave as no-operation instructions.  As such, programs must use global memory (in combination with procedural flow-control mechanisms) to correctly respond to environmental signals.

For each experimental condition, we evolved 200 replicate populations of 1000 programs for 10,000 generations at four levels of problem difficulty: $K=2, 4, 8$, and 16.
For each replicate, we randomly generated a unique tag for each environmental signal, and we initialized populations with randomly generated programs.
Each generation, we evaluated programs independently, and we selected programs using size-eight tournament selection.

\subsection{Contextual-signal Problem}
\label{sec:methods:contextual-signal-problem}

The contextual-signal problem is inspired by Skocelas and DeVries' method for verifying the functionality of recurrent neural network implementations \citep{skocelas_test_2020}. 
In the contextual-signal problem, programs must respond appropriately to a pair of input signals.
The order of these signals does not matter, but the first signal must be remembered (as ``context'') in order to produce the correct response to the second signal.
In this work, there are a total of four possible input signals and four possible outputs.
Programs output a particular response by executing one of four response instructions. 
Table \ref{tab:context-signal-input-combos} gives the correct output type for each pairing of input signals.

\renewcommand{\arraystretch}{1}
\begin{table}[ht!]
    \small
    \centering
    \begin{tabular}{c | c | c}
        \toprule
        Test case ID & Input Sequence & Correct Response \\ \hline 
        0 &
        S-0, S-0 &
        Response-A \\ 
        % \hline
        1 &
        S-0, S-1 &
        Response-B \\  
        % \hline
        2 &
        S-0, S-2 &
        Response-C \\  
        % \hline
        3 &
        S-0, S-3 &
        Response-D \\ 
        % \hline
        
        4 &
        S-1, S-0 &
        Response-B \\ 
        % \hline
        5 &
        S-1, S-1 &
        Response-C \\ 
        % \hline
        6 &
        S-1, S-2 &
        Response-D \\ 
        % \hline
        7 &
        S-1, S-3 &
        Response-A \\ 
        % \hline
        
        8 &
        S-2, S-0 &
        Response-C \\ 
        % \hline
        9 &
        S-2, S-1 &
        Response-D \\ 
        % \hline
        10 &
        S-2, S-2 &
        Response-A \\ 
        % \hline
        11 &
        S-2, S-3 &
        Response-B \\ 
        % \hline
        
        12 &
        S-3, S-0 &
        Response-D \\ 
        % \hline
        13 &
        S-3, S-1 &
        Response-A \\ 
        % \hline
        14 &
        S-3, S-2 &
        Response-B \\ 
        % \hline
        15 &
        S-3, S-3 &
        Response-C \\ 
        \bottomrule
    \end{tabular}
    \caption{\small 
    \textbf{Input signal sequences for the contextual-signal problem.} 
    }
    \label{tab:context-signal-input-combos}
\end{table} 

We evaluate programs on each of the 16 possible sequences of input signals (Table \ref{tab:context-signal-input-combos}); we consider each of these input sequences as a single test case.
For each test case evaluation, we give programs \ContextSigTaskCpuTimePerSignal\ time steps to process each signal.
After the first input signal, a program must update internal state information to ensure that the second input signal induces the correct response.  
Once the allotted time expires after the first input signal, the program's threads of execution are reset, resulting in a loss of all thread-local memory; \textit{only} the contents of global memory and each function's regulatory state persist. 
The program then receives the second input signal and must execute the correct response instruction within \ContextSigTaskCpuTimePerSignal\ time steps.
A program is considered a solution if it produces the correct response for all 16 possible sequences of input signals.

\subsubsection{Experimental Design}

We use the contextual-signal problem to
(1) assess the capacity of tag-based genetic regulation to perform context-dependent module execution based on \textit{distinct} input types
and
(2) evaluate the problem-solving success of a regulation-enabled GP system relative to an otherwise identical GP system with regulation disabled.
As in the signal-counting problem, we compared the problem-solving success of regulation-on and regulation-off GP systems.

For each experimental condition, we evolved 200 replicate populations of 1000 programs for 10,000 generations.
For each replicate, we randomly generated the tags associated with each type of input signal, and we initialized populations with randomly generated programs.
Instead of selecting programs to propagate based on an aggregate fitness measure, we used the lexicase parent selection algorithm \citep{helmuth_solving_2015} in which each combination of input signals (\textit{i.e}., row in Table \ref{tab:context-signal-input-combos}) constituted a single test case. 

\subsection{Boolean-logic Calculator Problem}
\label{sec:methods:boolean-calc-problem}
% alternative names
% - Bitwise Logic Calculator Problem

Inspired by Yeboah-Antwi's PushCalc system \citep{yeboah-antwi_evolving_2012}, 
the Boolean-logic calculator problem requires programs to implement a push-button calculator capable of performing each of the following 10 bitwise logic operations: ECHO, NOT, NAND, AND, OR-NOT, OR, AND-NOT, NOR, XOR, and EQUALS. 
Table \ref{tab:boolean-calc-operations} gives a brief overview of each of these operations. 
In this problem, there are 11 distinct types of input signals: one for each of the 10 possible operators and one for numeric inputs. 
Each distinct signal type is associated with a unique tag (randomly generated per-replicate) and is meant to recreate the context that must be maintained on a physical calculator.
Programs receive a sequence of input signals in prefix notation, starting with an operator signal and followed by the appropriate number of numeric input signals (that each contain an operand to use in the computation).
After receiving the appropriate input signals, programs must output the correct result of the requested computation.

% http://myxo.css.msu.edu/papers/nature2003/Supp2.html

\begin{table}[h!]
    \small
    \centering
    \begin{tabular}{c | c | c}
        \toprule
        Operation & \# Inputs & NAND gates \\ \hline 
        ECHO &
        1 &
        0
        \\
        NOT &
        1 &
        1
        \\
        NAND &
        2 &
        1
        \\
        AND &
        2 &
        2
        \\
        OR-NOT &
        2 &
        2
        \\
        OR &
        2 &
        3
        \\
        AND-NOT &
        2 &
        3
        \\
        NOR &
        2 &
        4
        \\
        XOR &
        2 &
        4
        \\
        EQUALS &
        2 &
        5
        \\
        \bottomrule
    \end{tabular}
    \caption{\small 
    \textbf{Bitwise Boolean logic operations used in the Boolean-logic calculator problem.} 
    Programs are given a \code{nand} instruction and must construct each of the other operations (aside from ECHO) out of \code{nand} operations.
    As such, we measure the difficulty of each operation as the minimum number of NAND gates required to construct the given operation.
    }
    \label{tab:boolean-calc-operations}
\end{table}

Programs are evaluated on a set of test cases (\textit{i.e.}, input/output examples) where each test case comprises a particular operator, the requisite number of operands, and the expected numeric output.
Test cases are evaluated on a pass/fail basis, and a program is classified as a solution if it passes all test cases in a training and testing set\footnote{
We use the testing set only to determine if a program can be categorized as a solution. The testing set is never used by the parent-selection algorithm to determine reproductive success.
}.
The training and testing sets used in this work are included in our supplemental material \citep{lalejini_supplement} and contained 442 and 5810 test cases, respectively.
Each generation, we sample 20 test cases from the training set, and we independently evaluate each program in the population on the sampled test cases.

When evaluating a program on a test case, we provide \CalculatorTaskCpuTimePerSignal\ time steps to process each input signal.
After time expires, the program's threads of execution are reset, resulting in a loss of all thread-local memory; only the contents of global memory and each function's regulatory state persist.
Because input signals are given in prefix notation, programs must adjust their internal state to ensure that the program performs and outputs the result of the appropriate computation after receiving the requisite number of operand input signals.

% # training cases: 442
% # testing cases: 5810
% # total cases: 6252
% Number of training examples to evaluate each generation: 20
%   Test case type: NOR; TypeID: 9; # training cases: 45; # training sample eval: 2
%   Test case type: ANDNOT; TypeID: 8; # training cases: 45; # training sample eval: 2
%   Test case type: XOR; TypeID: 7; # training cases: 45; # training sample eval: 2
%   Test case type: NOT; TypeID: 6; # training cases: 41; # training sample eval: 2
%   Test case type: ECHO; TypeID: 5; # training cases: 41; # training sample eval: 2
%   Test case type: EQU; TypeID: 1; # training cases: 45; # training sample eval: 2
%   Test case type: NAND; TypeID: 0; # training cases: 45; # training sample eval: 2
%   Test case type: ORNOT; TypeID: 2; # training cases: 45; # training sample eval: 2
%   Test case type: AND; TypeID: 3; # training cases: 45; # training sample eval: 2
%   Test case type: OR; TypeID: 4; # training cases: 45; # training sample eval: 2

\subsubsection{Experimental Design}

We use the Boolean-logic calculator problem to assess the utility of tag-based regulation on a challenging program synthesis problem. 
The signal-counting and contextual-signal problems each require programs to perform different computations in response to input signals, but those computations are abstracted as `response' instructions. 
The Boolean-logic calculator problem requires programs to both dynamically adjust which modules are executed in response to input signals and perform non-trivial computations on numeric inputs. 

We compared the problem-solving success of programs evolved in regulation-on and regulation-off conditions.
For each condition, we evolved 200 replicate populations of 1000 programs for 10,000 generations.
For each replicate, we randomly generated the tags associated with each type of input signal, and we initialized populations with randomly generated programs.
We selected parents using a variant of the down-sampled lexicase algorithm \citep{hernandez_random_2019}, guaranteeing that at least one of each type of test case (\textit{i.e.}, at least one of each type of operator) was used during evaluation. 

\subsection{Independent-signal Problem}
\label{sec:methods:independent-signal-problem}

% ----- BEGIN REVISION -----

The independent-signal problem requires programs to execute a unique response for each of 16 distinct input signals. 
Because signals are distinct, programs need not alter their response to any particular signal over time.
Instead, programs may ``hardwire'' each of the 16 possible responses to the appropriate input signal.
However, input signals are presented in a random order; thus, the correct \textit{order} of responses cannot be hardcoded.
Otherwise, evaluation (and fitness assignment) on the independent-signal task mirrors that of the signal-counting task (Section \ref{sec:methods:signal-counting-problem}).
A program is considered a solution if it responds correctly to all 16 input signals during evaluation.

\subsubsection{Experimental Design}

%%% @AML: working on highlighting how fitness eval is deliberately imperfect.
% - Problems of this nature are common: robot simulations, or large input spaces where only possible to test on small sample of input types. 
We deliberately configured fitness evaluation and solution identification in the independent-signal problem to be noisy and thus unreliable: each program is evaluated once on a single random ordering of input signals, and we label a program as a solution if it performs optimally during a single evaluation.
Because programs receive input signals in a random order, erroneous genetic regulation can manifest as cryptic variation (\textit{i.e.}, behavioral variation that is not expressed and selected on).
For example, non-adaptive down-regulation of a particular response function may be neutral given one sequence of input signals, but may be deleterious in another.
Indeed, this form of non-adaptive cryptic variation can also result from erroneous flow control structures. 

The independent-signal problem allows us to test whether genetic regulation can impede adaptive evolution in scenarios where outputs are not context-dependent and where fitness evaluation does not reliably differentiate between generalizing and non-generalizing candidate solutions.
Fitness evaluation for the independent-signal problem is computationally inexpensive, so we could easily increase the reliability of evaluation by testing programs on multiple orderings of input sequences.
However, our goal is not to demonstrate that we can \textit{solve} this diagnostic problem.
Rather, we aim to determine if this diagnostic represents a general scenario where unnecessary tag-based regulation can impede adaptive evolution relative to not having regulation. 

% While stochastic elements in fitness can impede problem-solving success and generalization, our goal here is to study how maladaptive regulation could further exacerbate these problems.
%, [as many real-world problems are likely to have unreliable evaluation (\textit{e.g.}, robotics simulations, [...]).]
% The changing-signal problem assesses the potential for unnecessary tag-based genetic regulation to \textit{impede} adaptive evolution relative to not having regulation.

As in each of the previous experiments, we compared programs evolved in regulation-on and regulation-off conditions.
Specifically, we compared initial prob\-lem-solving success and how well solutions generalized to a sample of 5000 input sequences (of ${\sim}2.1\times10^{13}$ possible sequences). 
We deemed programs as having generalized only if they responded correctly in all 5000 tests.

We evolved 200 replicate populations of 1000 programs for 10,000 generations under each condition. 
For each replicate, we randomly generated 16 unique input signal tags.
All other experimental procedures were identical to that of the signal-counting task.

% ------ END REVISION ------

\subsection{Data Analysis and Reproducibility}

% How knockout experiments work
For each replicate in a given experiment, we extracted and analyzed the first evolved program that was classified as a solution. 
We compared the number of successful replicates (\textit{i.e.}, replicates that yielded a solution) across experimental conditions using Fisher's exact test.
%, and we applied a Bonferroni correction for multiple comparisons where appropriate. %@AML: no actual multiple comparisons in results!
We conducted knockout experiments on successful programs to identify the mechanisms underlying their behavior.
In all knockout experiments, we re-evaluated programs with a target functionality (\textit{e.g.}, regulation instructions) replaced with no-operation instructions.
Specifically, we independently knocked out (1) all regulatory instructions, (2) all instructions that access a program's global memory buffer, and (3) both regulatory instructions and global memory access instructions. 
We classify a program as reliant on a particular functionality if, when knocked out, fitness decreases. 
In addition to knockout experiments, we tracked the distribution of instruction types (\textit{e.g.}, flow control, mathematical operations, \textit{etc.}) executed by successful programs. 
For each successful replicate, we extracted the proportion of flow control instructions (\textit{i.e.}, conditional logic instructions such as ``if'' or ``while'' statements) executed by the evolved solution. 
We compared the proportions of flow control instructions executed by regulation-on solutions and regulation-off solutions, allowing us to assess the relative importance of conditional logic across experimental treatments.

For programs reliant on genetic regulation, we abstracted regulatory networks as directed graphs by monitoring program execution.
Vertices represent program functions, and directed edges (each categorized as promoting or repressing) show the regulatory interactions between two functions.
For example, a repressing edge from function A to function B indicates that B was repressed when A was executing.

We implemented our experiments using the Empirical scientific software library \citep{charles_ofria_2020_empirical}, and we conducted all statistical analyses using R version 4.0.2 \citep{r_language_2020}.
We used the reshape2 \cite{R-reshape2} R package and the tidyverse \citep{r_tidyverse_2019} collection of R packages to wrangle data. 
We used the following R packages for graphing and visualization: ggplot2 \citep{R-ggplot2}, cowplot \citep{R-cowplot}, viridis \citep{R-viridis}, Color Brewer \citep{harrower_colorbrewerorg_2003,R-Brewer_2014}, and igraph \citep{igraph2006}.
We used R markdown \citep{rmarkdown} and bookdown \citep{R-bookdown} to generate web-enabled supplemental material.
Our source code for experiments and analyses, along with guides for replication, can be found in supplemental material \citep{lalejini_supplement}, which is hosted on \href{https://github.com/amlalejini/Tag-based-Genetic-Regulation-for-LinearGP/}{GitHub}.
Additionally, we have made all of our experimental data available on the \href{https://osf.io/928fx/}{Open Science Framework} (see \supSecDataAvailability\ in supplement \citep{lalejini_supplement}).

\section{Results and Discussion}

\subsection{Tag-based regulation improves problem-solving performance on context-dependent tasks}

% Overview paragraph (outline results across all three context-dependent problems)
We found that tag-based regulation improves performance on each of the three problems that require context-dependent behavior: the signal-counting problem (Section \ref{sec:results:signal-counting-problem}), contextual-signal problem (Section \ref{sec:results:context-signal-problem}), and Boolean-logic calculator problem (Section \ref{sec:results:boolean-calc-problem}). 
Additionally, we conducted knockout experiments that confirmed that evolved tag-based regulation allows solutions to dynamically adjust module execution over time. 
We also found that, across all three context-dependent problems, regulation-off solutions (\textit{i.e.}, solutions evolved using regulation-disabled SignalGP) executed a larger proportion of conditional logic instructions than regulation-on solutions (\textit{i.e.}, solutions evolved using regulation-enabled SignalGP). 
This result suggests that without regulation, programs must evolve larger, more complex conditional logic structures. 

\subsubsection{Signal-counting Problem}
\label{sec:results:signal-counting-problem}

\begin{table}[h!]
    \small
    \centering
    \begin{tabularx}{\columnwidth}{cYY}
        \toprule
          & Regulation-off condition 
          & Regulation-on condition \\
        %  \hline
        \cmidrule(lr){1-3}
         Two-signal   
            & \RSPTwoSigMemOnlySolutionCnt  % Memory-only
            & \RSPTwoSigBothSolutionCnt \\  % Both
        %  \hline
         Four-signal
            & \RSPFourSigMemOnlySolutionCnt  % Memory-only
            & \RSPFourSigBothSolutionCnt \\  % Both
        %  \hline
         Eight-signal 
            & \RSPEightSigMemOnlySolutionCnt  % Memory-only
            & \RSPEightSigBothSolutionCnt \\  % Both
        %  \hline
         Sixteen-signal
            & \RSPSixteenSigMemOnlySolutionCnt  % Memory-only
            & \RSPSixteenSigBothSolutionCnt \\  % Both
         \bottomrule
    \end{tabularx}
    \caption{\small
    \textbf{Signal-counting problem-solving success.}
    This table gives the number of successful replicates (\textit{i.e.}, in which a perfect solution evolved) out of 200 on the signal-counting problem across four problem difficulties and two experimental conditions. 
    For each problem difficulty, the regulation-off condition was less successful than the regulation-on condition (Fisher's exact test; all difficulties: $p < 10^{-15}$).
    }
    \label{tab:signal-counting-solutions}
\end{table}

% two-signal: p-value < 2.2e-16, 
% four-signal: p-value < 2.2e-16
% eight-signal: p-value < 2.2e-16
% sixteen-signal: p-value < 2.2e-16

\begin{figure}[htbp]
    \centering
    \includegraphics[width=0.7\textwidth]{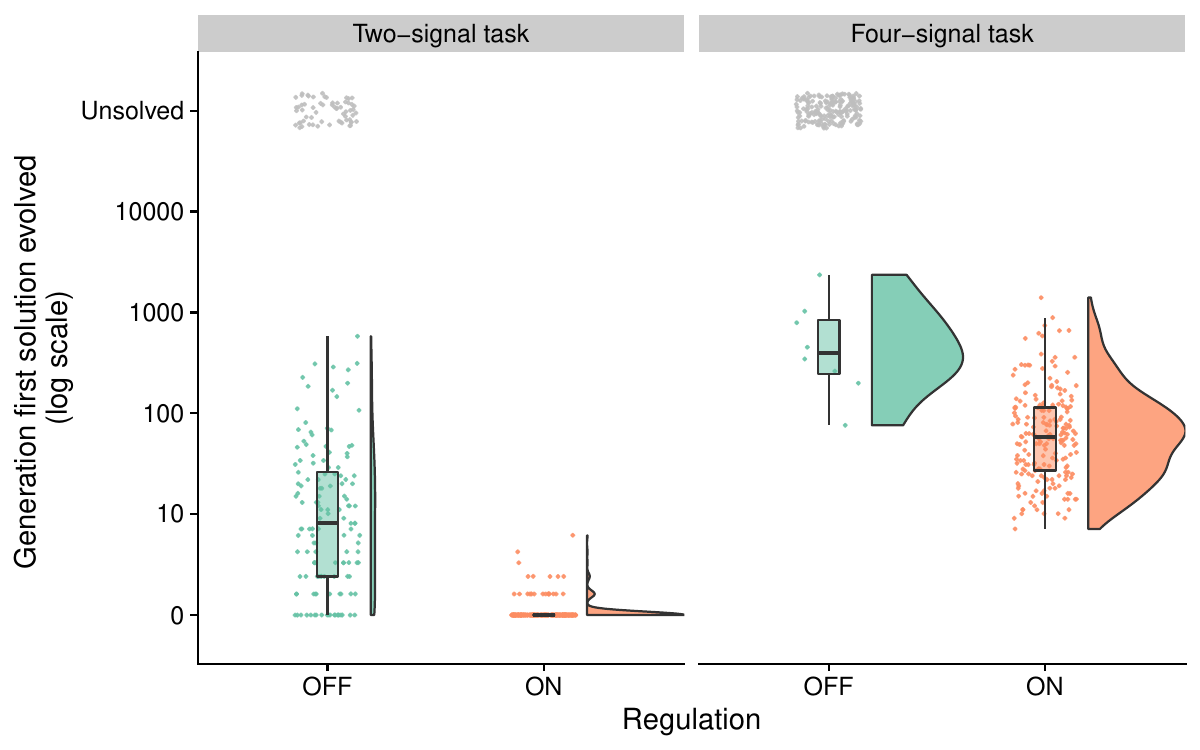}
    \caption{\small
    \textbf{Generation at which first solution evolved (log scale) in each successful replicate for the signal-counting problem (Raincloud plot \citep{allen_raincloud_2019}).}
    We show data from only those problem difficulties in which solutions evolved (two- and four-signal problems).
    Gray points indicate the number of unsuccessful replicates for each condition.
    For both problem difficulties, regulation-on solutions typically required fewer generations than regulation-off solutions to arise (Wilcoxon rank sum test; two-signal: $p < 10^{-15}$, four-signal: $p < 9\times10^{-05}$). 
    }
    \label{fig:signal-counting-solve-time}
\end{figure}

% (1) # Solution & solution timing
Table \ref{tab:signal-counting-solutions} shows the results from the signal-counting problem for each experimental condition across all four levels of problem difficulty. 
Regulation-on conditions consistently yielded a larger number of successful replicates than regulation-off conditions where programs relied on their global memory buffer in combination with procedural flow control for success.
Although global memory is technically sufficient to solve each version of the signal-counting problem\footnote{
We verified this claim by hand-coding solutions that rely on global memory and flow-control instructions (supplemental \supSecHandcodedPrograms\ \citep{lalejini_supplement}).
}, 
in practice such solutions evolved in only the two- and four-signal variants.
Tag-based regulation, in contrast, appears more readily adaptive, as regulation-based solutions arose across all problem difficulties, implying that access to tag-based regulation can drive increased problem-solving success.
Further, we found that, in the two- and four-signal tasks, solutions arose after significantly fewer generations in the regulation-on conditions than in the regulation-off controls (Figure \ref{fig:signal-counting-solve-time}).

% two-signal - p-value < 2.2e-16
% four-signal 8.603e-05

% --- BEGIN REVISION ---
Tag-based regulation renders the two-signal task trivial: all solutions evolved in under 10 generations. 
In fact, the majority of regulation-on solutions (178 out of 200) were found in the initial randomly generated population. 
However, not all replicates without access to tag-based regulation even found a solution to the two-signal task.
% ---- END REVISION ---

\begin{table}[htbp]
    \small
    \centering
    \begin{tabularx}{\columnwidth}{cYYY}
        \toprule
          & No regulation required
          & Regulation required 
          & Unsolved \\
        %  \hline
        \cmidrule(lr){1-4}
         Two-signal     
            & \RSPTwoSigMemOnlyStrategyCnt  % Memory-only
            & \RSPTwoSigRegRequiredStrategyCnt  
            & \RSPTwoSigUnsolvedCnt \\
        %  \hline
         Four-signal    
            & \RSPFourSigMemOnlyStrategyCnt  % Memory-only
            & \RSPFourSigRegRequiredStrategyCnt   % 
            & \RSPFourSigUnsolvedCnt \ \\
        %  \hline
         Eight-signal   
            & \RSPEightSigMemOnlyStrategyCnt  % Memory-only
            & \RSPEightSigRegRequiredStrategyCnt  % 
            & \RSPEightSigUnsolvedCnt \\
        %  \hline
         Sixteen-signal 
            & \RSPSixteenSigMemOnlyStrategyCnt  % Memory-only
            & \RSPSixteenSigRegRequiredStrategyCnt  % 
            & \RSPSixteenSigUnsolvedCnt \\
         \bottomrule
    \end{tabularx}
    \caption{\small 
    \textbf{Mechanisms underlying solutions from the regulation-on condition for the signal-counting problem.}
    To determine a successful program's underlying strategy, we re-evaluated the program with global memory access instructions knocked out (\textit{i.e.}, replaced with no-operation instructions) and with regulation instructions knocked out.
    This table shows the number of regulation-on solutions that actually rely on regulation to solve the signal-counting problem. 
    }
    \label{tab:signal-counting-ko-strategies}
\end{table} 

We conducted knockout experiments to investigate the mechanisms underlying successful programs. 
Indeed, all solutions evolved without access to tag-based regulation relied exclusively on their global memory buffer to differentiate their behavior (see supplemental \supSecRepeatedSigAnalysis\ \citep{lalejini_supplement}). 
Table \ref{tab:signal-counting-ko-strategies} shows the strategies used by programs evolved with regulation-enabled SignalGP.
Our knockout experiments confirm that the majority of solutions evolved with access to tag-based regulation do indeed rely on regulation to dynamically adjust their responses to signals over time.

\begin{figure}[ht]
\centering

\begin{subfigure}[b]{0.55\textwidth}
    \centering
    \includegraphics[width=\linewidth]{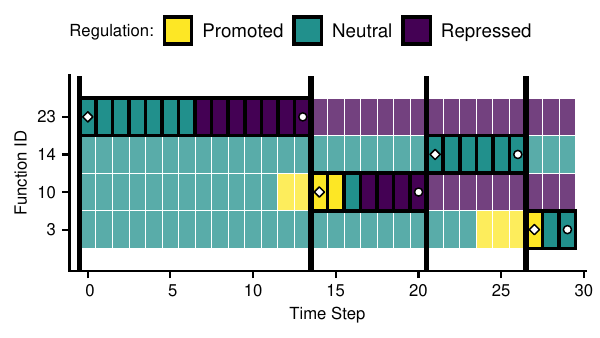}
    \caption{\small Module regulation over time.}
    \label{subfig:rst-exec-trace}
\end{subfigure}%
\hfill
\begin{subfigure}[b]{0.3\textwidth}
    \centering
    \includegraphics[width=\linewidth]{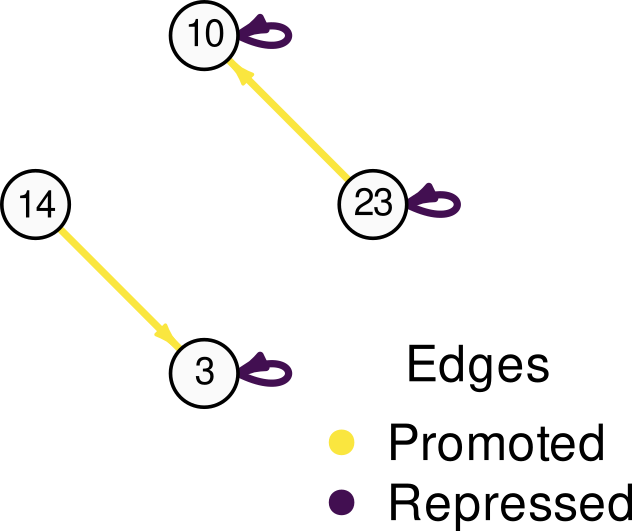}
    \caption{\small Regulatory network.}
    \label{subfig:rst-reg-network}
\end{subfigure}%

\caption{\small 
   \textbf{Execution trace of a SignalGP program solving the four-signal version of the signal-counting task.}
    Color denotes each function's regulatory state (yellow: promoted, purple: repressed) during evaluation; functions not regulated or executed are omitted.
    Functions that are actively executing are annotated with a black outline.
    Black vertical lines denote input signals, and a diamond (white with black outline) indicates which function was triggered by the input signal.
    A circle (white with black outline) indicates which function executed a response.
    (b) shows the directed graph representing the regulatory network associated with trace (a).
    Vertices depict functions that either ran during evaluation or were regulated. 
    Each directed edge shows a regulatory relationship between two functions where the edge's source acted on (promoted in yellow or repressed in purple) the edge's destination.
    Note that in the case presented here all repressing relationships are self-referential.
    }
    
\label{fig:signal-counting-example-networks}
\end{figure}

% (3) Evolved networks
We further assessed the functionality of tag-based regulation by analyzing the execution traces of evolved solutions.
We visualized the gene regulatory networks that manifest as a result of programs executing promoter and repressor instructions. 
Figure \ref{fig:signal-counting-example-networks} overviews the execution of a representative evolved program on the four-signal instance of the signal-counting problem.
We found that successful programs tend to operate via a succession of self-repressing events where modules express the appropriate response then disable themselves so that the next best-matching function---expressing the appropriate next response---will activate instead.
This behavioral pattern continues for each subsequent environmental signal.
Indeed, across all problem difficulties, we observed that successful regulatory networks generally contained more repression relationships than promotion relationships between functions (supplemental \supSecRepeatedSigAnalysis\  \citep{lalejini_supplement}).
Independent knockouts of up-regulation and down-regulation confirm that the majority of successful regulatory networks rely on down-regulation: 
of the 661 successful regulatory networks evolved across all problem difficulties, 392 rely exclusively on down-regulation, 7 rely exclusively on up-regulation, 259 rely on both up- and down-regulation, and 3 rely on \textit{either} up- or down-regulation (\textit{i.e.}, they required regulation but were robust to independent knockouts of up- and down-regulation). 

Our experimental data highlights the benefit of tag-based genetic regulation in addition to traditional, register-based means of dynamically adjusting responses to a repeated input signal over time. 
However, our data may also indicate a deficiency in the design of SignalGP's current global memory model.
An improved memory model may also enhance the capacity for programs to dynamically adjust their responses to inputs over time; however, any memory-based solution will still suffer from the need to incorporate flow-control structures to implement this functionality, inherently creating a larger evolutionary hurdle to overcome.
Indeed, we found that the memory-based solutions that evolved in our experiments executed a larger proportion of flow-control instructions than regulation-based solutions
(Wilcoxon rank sum test; two-signal: $p < 10^{-10}$, four-signal: $p = 0.004$; supplemental \supSecRepeatedSigAnalysis\ \citep{lalejini_supplement}).
% two-signal: p-value = 2.118e-11; 95% ci 2% to 3.7% difference in proportion
% four-signal: p-value = 0.003185; 95% ci 1.56% to 6.4% difference in proportion

\subsubsection{Contextual-signal Problem}
\label{sec:results:context-signal-problem}

% (1) Solutions + solution timings

\begin{figure}[ht]
\centering

\begin{subfigure}[b]{0.45\textwidth}
    \centering
    \includegraphics[width=\linewidth]{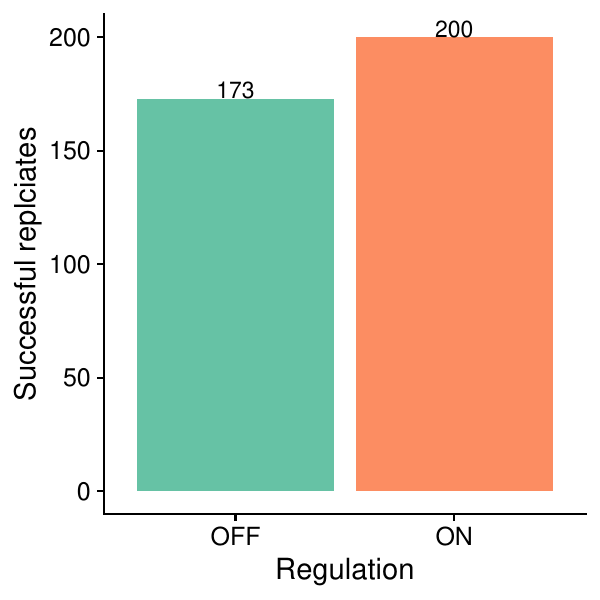}
    \caption{\small Successful replicates.}
    \label{subfig:context-signal-solution-counts}
\end{subfigure}
\hfill
\begin{subfigure}[b]{0.45\textwidth}
    \centering
    \includegraphics[width=\textwidth]{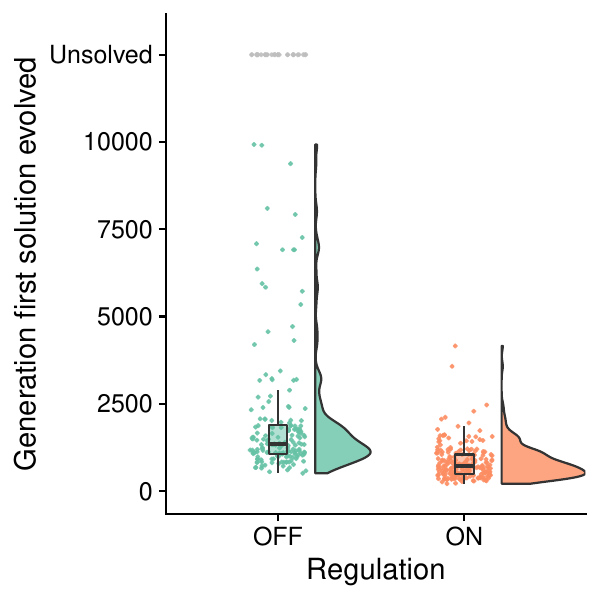}
    \caption{\small Generations elapsed before solution.}
    \label{subfig:context-signal-solve-time}
\end{subfigure}

\caption{\small 
\textbf{Contextual-signal problem-solving performance.}
(a) shows the number of successful replicates for the regulation-off and regulation-on conditions on the contextual-signal problem. 
The regulation-off condition was less successful than the regulation-on condition (Fisher's exact test: $p < 6\times10^{-9}$).
(b) is a Raincloud plot showing the generation at which the first solution evolved in each successful replicate.
Gray points indicate the number of unsuccessful replicates for each condition.
Regulation-on solutions typically required fewer generations than regulation-off solutions to arise  (Wilcoxon rank sum test: $p < 10^{-15}$).
}

% problem-solving success - p-value = 5.818e-09; fisher's
% solve time: p-value < 2.2e-16; wilcoxon
    
\label{fig:context-signal-performance}
\end{figure}

Figure \ref{subfig:context-signal-solution-counts} shows the number of successful replicates on the contextual-signal problem for both the regulation-on and regulation-off conditions.
While both conditions were often successful, we found that access to tag-based regulation significantly improved problem-solving success.
Further, regulation-on solutions typically required fewer generations to evolve than regulation-off solutions (Figure \ref{subfig:context-signal-solve-time}).

% (2) Solution strategies (knockouts)
We used knockout experiments to identify the mechanisms underlying each solution's strategy.
As expected, all 173 solutions evolved without access to tag-based regulation relied on their global memory buffer to track contextual information and used control flow mechanisms to differentiate their responses based on stored context.
Indeed, we found that regulation-off solutions executed a larger proportion of flow-control instructions than regulation-on solutions (Wilcoxon rank sum test: $p < 10^{-15}$; supplement \supSecContextSigAnalysis\ \citep{lalejini_supplement}).
We also found that all 200 regulation-on solutions relied on tag-based regulation for response differentiation: 105 relied only on tag-based regulation and 95 relied on a combination of both tag-based regulation and global memory.
% Control flow proportions:
% p < p-value < 2.2e-16; 95% ci: 4.28% - 5.43%

In contrast to the signal-counting problem, we did not find that successful regulatory networks used primarily self-repressing modules.
Instead, we found that networks were more balanced between repressing and promoting edges; indeed, we found that successful networks generally contained more promoting edges than repressing edges (supplement \supSecContextSigAnalysis\ \citep{lalejini_supplement}).
This result suggests that we should expect different problems to select for different forms of gene regulatory networks.
% @AML: I assume there are some AGRN results that support that ^^

\subsubsection{Boolean-logic Calculator Problem}
\label{sec:results:boolean-calc-problem}

% (1) solutions + solution timings

\begin{figure}[ht]
\centering

\begin{subfigure}[b]{0.45\textwidth}
    \centering
    \includegraphics[width=\linewidth]{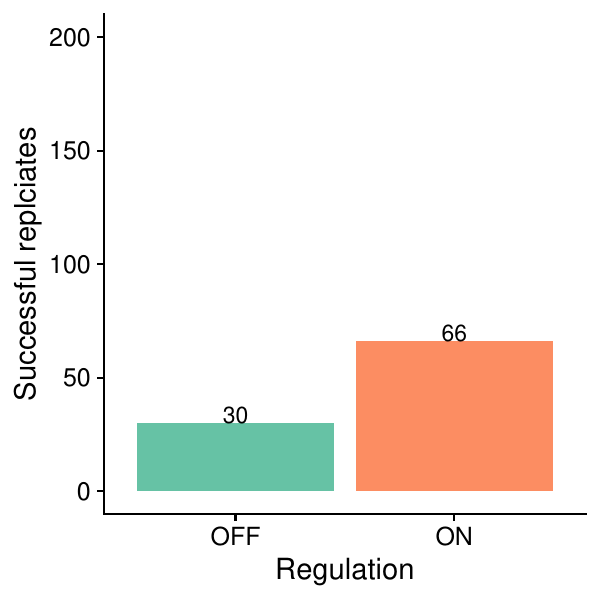}
    \caption{\small Successful replicates.}
    \label{subfig:boolean-calc-prefix-solution-count}
\end{subfigure}
\hfill
\begin{subfigure}[b]{0.45\textwidth}
    \centering
    \includegraphics[width=\textwidth]{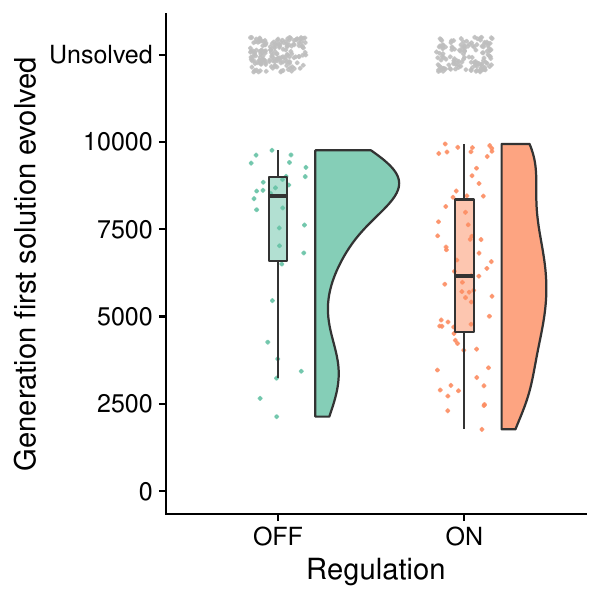}
    \caption{\small Generations elapsed before solution.}
    \label{subfig:boolean-calc-prefix-solve-time}
\end{subfigure}

\caption{\small
\textbf{Boolean-logic calculator problem-solving performance.}
(a) shows the number of successful replicates for the regulation-off and regulation-on conditions on the Boolean-logic calculator problem. 
The regulation-off condition was less successful than the regulation-on condition (Fisher's exact test: $p < 4\times10^{-05}$).
(b) is a Raincloud plot showing the generation at which the first solution evolved in each successful replicate.
Gray points indicate the number of unsuccessful replicates for each condition.
Regulation-on solutions typically required fewer generations than regulation-off solutions to arise (Wilcoxon rank sum test: $p < 0.042$).
}

% fisher's  p-value = 3.585e-05
% wilcoxon rank sum: p-value = 0.04102
    
\label{fig:boolean-calc-prefix-performance}
\end{figure}

Figure \ref{subfig:boolean-calc-prefix-solution-count} shows the number of successful replicates on the Boolean-logic calculator problem for both the regulation-on and regulation-off conditions.
While both regulation-on and regulation-off solutions evolved, we again found that access to genetic regulation significantly improved problem-solving success.
Further, as in the signal-counting and contextual-signal problems, regulation-on solutions typically required fewer generations to evolve than regulation-off solutions (Figure~\ref{subfig:boolean-calc-prefix-solve-time}).

\begin{figure}
\centering

\begin{subfigure}[b]{0.95\textwidth}
    \centering
    \includegraphics[width=\textwidth]{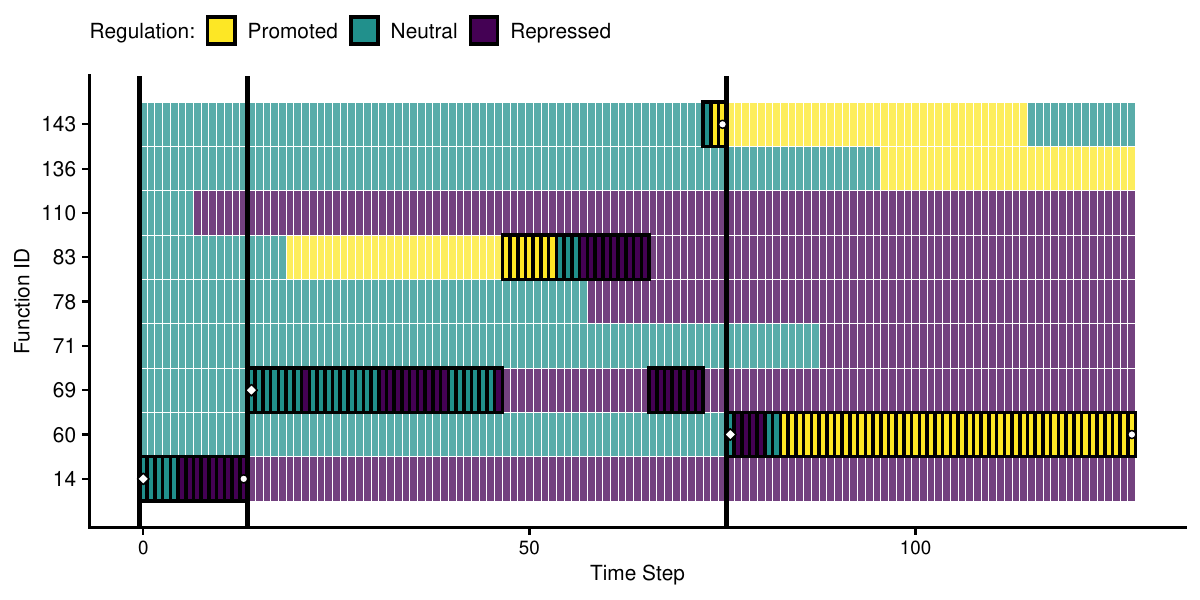}
    \caption{\small  Module regulation over time for a NAND operation.}
    \label{subfig:bc-nand-exec-trace}
\end{subfigure}%

\begin{subfigure}[b]{0.5\textwidth}
    \centering
    \includegraphics[width=0.6\textwidth]{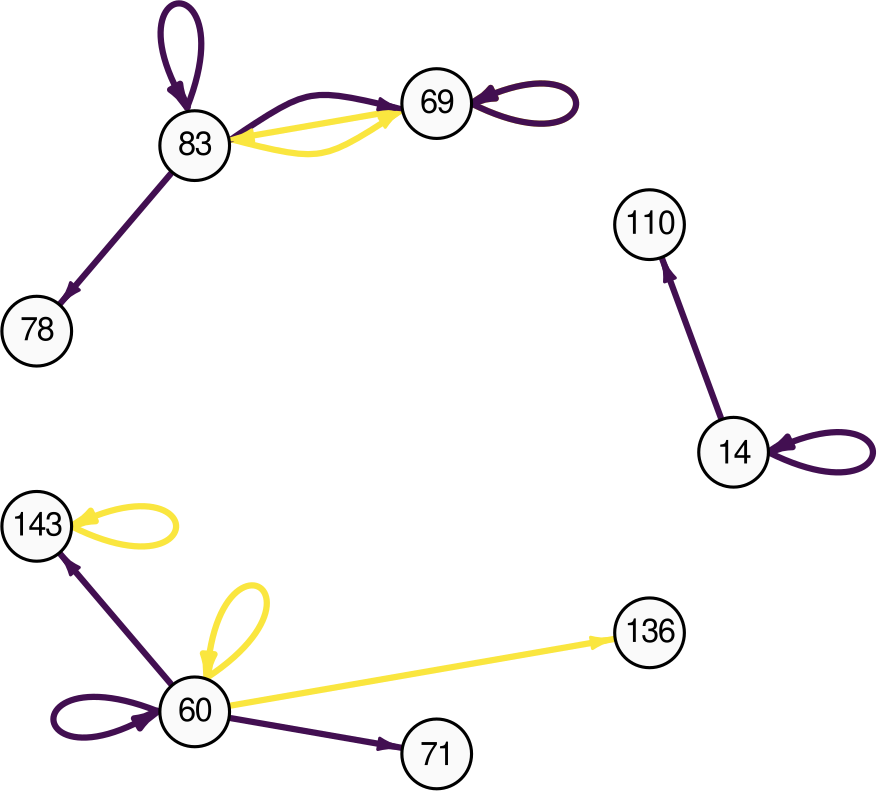}
    \caption{\small NAND regulatory network.}
    \label{subfig:bc-nand-reg-network}
\end{subfigure}%
\hfill
\begin{subfigure}[b]{0.5\textwidth}
    \centering
    \includegraphics[width=0.6\textwidth]{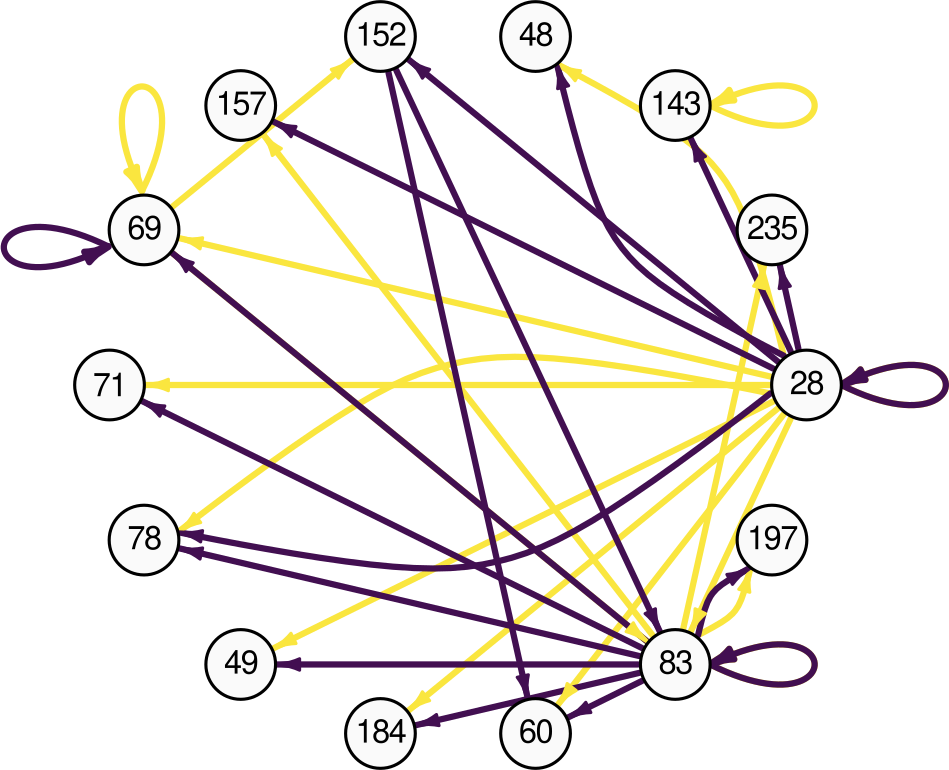}
    \caption{\small NOR regulatory network.}
    \label{subfig:bc-nor-reg-network}
\end{subfigure}%

\begin{subfigure}[b]{0.95\textwidth}
    \centering
    \includegraphics[width=0.95\textwidth]{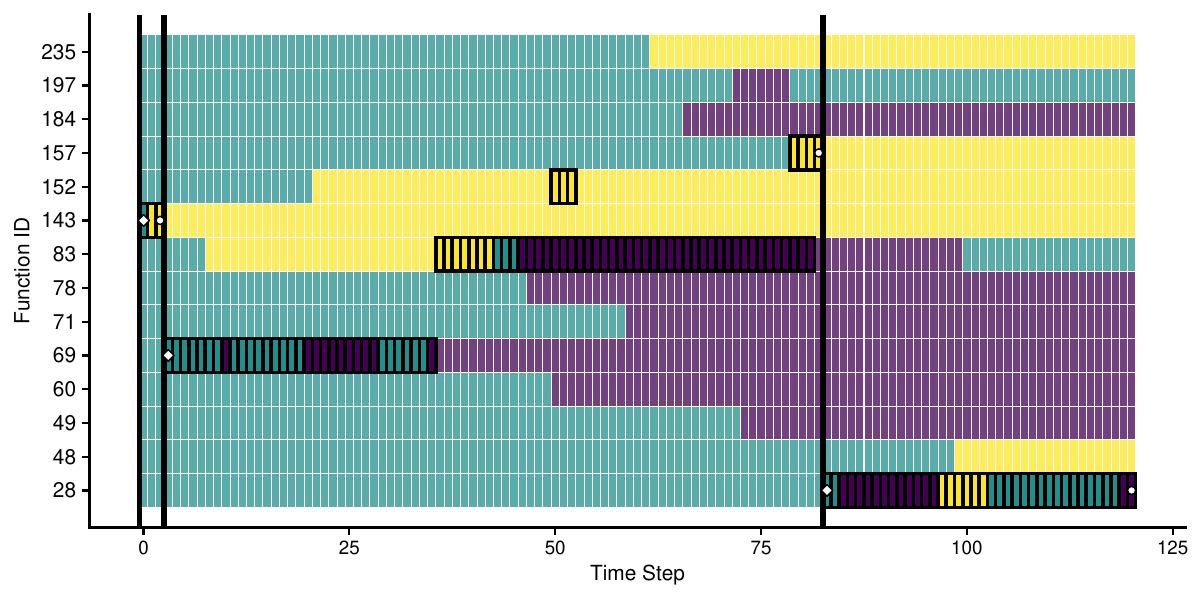}
    \caption{\small Module regulation over time for a NOR operation.}
    \label{subfig:bc-nor-exec-trace}
\end{subfigure}%

\caption{\small 
\textbf{Execution traces of a successful SignalGP program computing a NAND operation (a) and a NOR operation (d).}
(b) and (c) show the directed graphs representing the regulatory networks associated with traces (a) and (d), respectively.
These visualizations are in the same format as those in Figure \ref{fig:signal-counting-example-networks}.}
\label{fig:boolean-calc-prefix-example-networks}
\end{figure}

As in previous experiments, we conducted knockout experiments to identify the mechanisms underlying each solution's strategy.
To compute any of the Boolean logic operations, programs \textit{must} make use of the global memory buffer to store numeric inputs (operands) to be used when performing the computation specified by the final operator signal. 
Indeed, all solutions evolved across all conditions relied on their global memory buffer to solve this problem.
All 66 regulation-on solutions, however, also relied on tag-based regulation to perform the appropriate computation for each test case.
Consistent with results from each other context-dependent problem, we found that regulation-off solutions executed a larger proportion of flow-control instructions than regulation-on solutions (Wilcoxon rank sum test: $p < 2\times10^{-05}$; supplement \supSecBooleanCalcPrefixAnalysis\ \citep{lalejini_supplement}).
% Wilcoxon rank sum test: p-value = 1.328e-05 CI 1.486% - 3.91%

As in the signal-counting problem, we visualized the gene regulatory networks that manifest as a result of programs executing promoting and repressing instructions. 
Figure \ref{fig:boolean-calc-prefix-example-networks} overviews the execution of a representative program evolved to solve the Boolean-logic calculator problem.
Specifically, Figure \ref{fig:boolean-calc-prefix-example-networks} shows a program computing NAND and the same program computing NOR. 
The networks expressed on each of these operations are distinct despite originating from the same code.
These visualizations confirm that tag-based regulation allows programs to dynamically adjust their responses based on context (in this case, an initial operator signal). 

\subsection{Erroneous regulation can hinder task generalization}

In the signal-counting, contextual-signal, and Boolean-logic calculator problems, programs must adjust their behavior depending on the particular sequence of received signals.
The independent-signal problem, however, requires no signal-response plasticity; programs maximize fitness by statically associating $K$ distinct responses each with one of $K$ distinct input signals.
For this task, re-wiring signal-response associations within-lifetime is maladaptive.
As such, does the capacity for regulation impede adaptation to the independent-signal task?

We compared 200 replicate populations evolved with regulation-enabled SignalGP (``regulation-on'') and 200 populations evolved with regulation-disabled SignalGP (``regulation-off'').
All replicates produced a SignalGP program capable of achieving a perfect score during evaluation. 
We found no evidence that the availability of regulation affected the number of generations required to produce these solutions.

Next, we investigated how well evolved solutions \textit{generalized} across random permutations of input sequences.
Selection was deliberately based on a single stochastic ordering of environmental signals, so a ``perfect'' score may not generalize across all signal orderings.
% Because there are $K!$ possible sequences, each generation, we evaluated programs against random permutations of input signals.
% To test for generalization, we sampled 5000 input sequences (of ${\sim}2.1\times10^{13}$ possible sequences);
% we deem programs as having generalized only if they respond correctly in all 5000 tests.
We expect that programs evolved with access to regulation will more often exhibit non-adaptive plasticity that hinders generalization.

\begin{figure}[t]
    % \vspace{-10pt}
    \centering
    \includegraphics[width=0.39\textwidth]{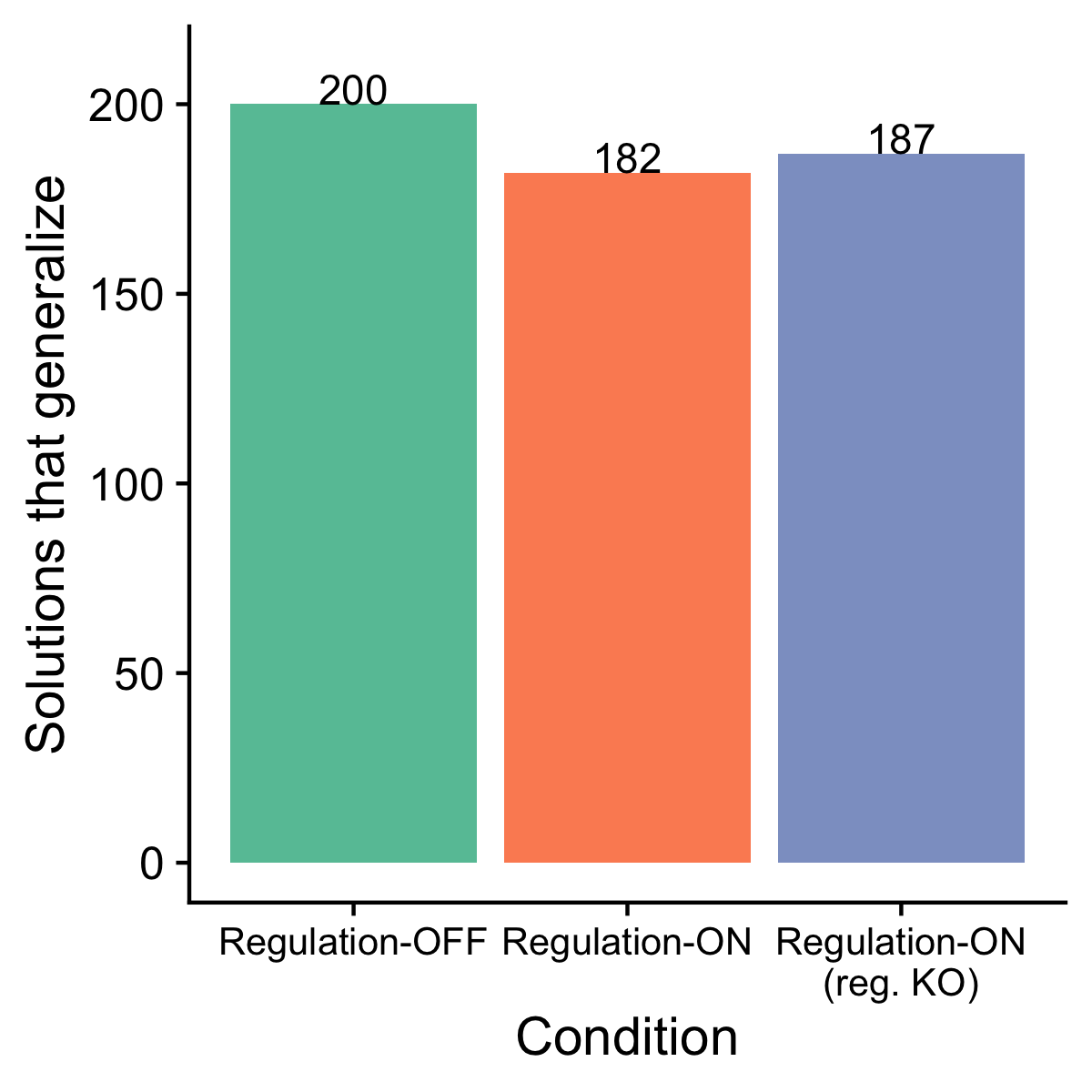}
    \caption{\small
    \textbf{The number of evolved solutions that generalize on the independent-signal problem.}
    The difference in number of solutions that generalize between the regulation-on and regulation-off conditions is statistically significant (Fisher's exact test: $p < 6\times10^{-06}$).
    The ``Regulation-ON (reg. KO)'' condition comprises the solutions from the Regulation-on condition, except with regulatory instructions knocked out (\textit{i.e.}, replaced with no-operation instructions).
    }
    % \vspace{-10pt}
    \label{fig:independent-signal-generalization}
\end{figure}

Figure \ref{fig:independent-signal-generalization} shows the number of evolved solutions from each condition that successfully generalized.
All programs that evolved without access to regulation successfully generalized; however, evolved programs from 18 out of 200 successful regulation-on replicates failed to generalize beyond the test cases they experienced during evolution (Fisher's exact test: $p < 6\times10^{-6}$).
Moreover, 5 of 18 non-generalizing programs generalized when we knocked out tag-based regulation.
Upon closer inspection, the other non-general programs relied on tag-based regulation for initial success but failed to generalize to arbitrary environment sequences.
% fisher's p-value = 5.113e-06

% \begin{wrapfigure}{r}{0.4\textwidth}
%     \vspace{-10pt}
%     \centering
%     \includegraphics[width=0.39\textwidth]{media/chg-env-16-generalization.png}
%     \caption{\small
%     \textbf{The number of evolved solutions that generalize on the independent-signal problem.}
%     The difference in number of solutions that generalize between the regulation-on and regulation-off conditions is statistically significant (Fisher's exact test: $p < 6\times10^{-06}$).
%     The ``Regulation-ON (reg. KO)'' condition comprises the solutions from the Regulation-on condition, except with regulatory instructions knocked out (\textit{i.e.}, replaced with no-operation instructions).
%     }
%     % \vspace{-10pt}
%     \label{fig:independent-signal-generalization}
% \end{wrapfigure}

% --- BEGIN REVISION ---
Unexpressed traits that vary in a population (but do not affect fitness) are collectively known as cryptic variation. 
Cryptic variation is pervasive in nature and thought to play an important role in evolution, potentially acting as a cache of diverse phenotypic effects in novel environments \citep{gibson_uncovering_2004,paaby_cryptic_2014}.
% @AML: new line, could be stronger
Such cryptic variation has been shown to help GP systems escape local optima, improving overall problem-solving performance \citep{turner_neutral_2015}.
Cryptic variation arises when environmental conditions that would reveal the variation are not experienced. 
Access to tag-based regulation appears to make such cryptic variation in evolving programs a stronger possibility than previously.
This dynamic can be valuable for performing more realistic studies of evolutionary dynamics with digital organisms (\textit{i.e.}, self-replicating computer programs \citep{wilke_biology_2002}).
However, when using regulation-enabled SignalGP in problem-solving domains, such as automatic program synthesis, non-adaptive plasticity should be accounted for in fitness objectives.
In the independent-signal problem, for example, we could have performed more thorough evaluations of programs using multiple random permutations of input sequences instead of one.
In more challenging problems, however, more thorough evaluations can come at the cost of substantial computational effort. 
% --- END REVISION ---

\subsection{Reducing the context required for the Boolean-logic calculator problem eliminates the benefit of regulation}

Experimental results on the independent-signal problem suggest that enabling tag-based regulation is not necessarily beneficial for solving problems that do not require context-dependent responses to input. 
We use a modified version of the Boolean-logic calculator problem to further investigate the potential for tag-based regulation to impede adaptive evolution. 
The Boolean-logic calculator problem as described in Section \ref{sec:methods:boolean-calc-problem} provides inputs in prefix notation: the operator (\textit{e.g.}, AND, OR, XOR, \textit{etc.}) is specified first, followed by the requisite number of numeric operands. 
As such, the final input signal does not differentiate which type of computation a program is expected to perform. % (\textit{e.g.}, AND, OR, XOR, \textit{etc.}).
Programs must adjust their response based on the context provided by previous signals, thereby increasing the utility of regulation. 

Here, we explore whether the calculator problem's context-dependence is driving the benefit of tag-based regulation that we identified in Section \ref{sec:results:boolean-calc-problem}.
We can reduce context-dependence of the calculator problem by presenting input sequences in \textit{postfix} notation. 
In postfix notation, programs receive the requisite numeric operand inputs first and the operator input last. 
As such, the final signal in an input sequence will always differentiate which bitwise operation should be performed.
Successful programs must store the numeric inputs embedded in operand signals, and then, as in the independent-signal problem, a distinct signal will differentiate which of the response types a program should execute. 

\begin{figure}[ht]
\centering

\begin{subfigure}[b]{0.45\textwidth}
    \centering
    \includegraphics[width=\linewidth]{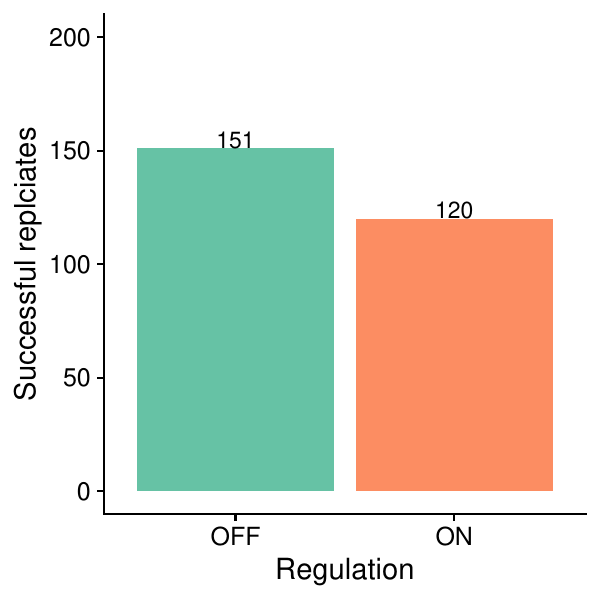}
    \caption{\small Successful replicates.}
    \label{subfig:boolean-calc-postfix-solution-count}
\end{subfigure}
\hfill
\begin{subfigure}[b]{0.45\textwidth}
    \centering
    \includegraphics[width=\textwidth]{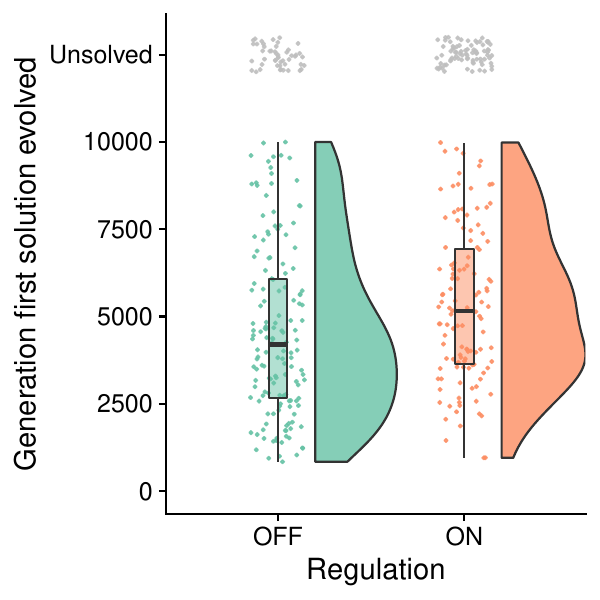}
    \caption{\small Generations elapsed before solution.}
    \label{subfig:boolean-calc-postfix-solve-time}
\end{subfigure}

\caption{\small 
\textbf{Boolean-logic calculator (postfix notation)  problem-solving performance.}
(a) shows the number of successful replicates for the regulation-off and regulation-on conditions on the postfix Boolean-logic calculator problem. 
The regulation-on condition was less successful than the regulation-off condition (Fisher's exact test: $p<0.002$).
(b) is a Raincloud plot showing the generation at which the first solution evolved in each successful replicate.
Gray points indicate the unsuccessful replicates for each condition.
Regulation-off solutions typically required fewer generations than regulation-on solutions to arise (Wilcoxon rank sum test: $p<0.004$).
}
% fisher's exact: p-value = 0.001286
% wilcoxon:  p-value = 0.003285
\label{fig:boolean-calc-postfix-performance}
\end{figure}

We repeated the Boolean-logic calculator experiment (as described in Section \ref{sec:methods:boolean-calc-problem}), except we presented inputs in postfix notation instead of prefix notation. 
Figure \ref{subfig:boolean-calc-postfix-solution-count} shows the number of successful replicates evolved in regulation-on and regulation-off conditions.
Postfix notation decreases the overall difficulty of the Boolean-logic calculator problem; more solutions evolved in each condition than evolved with prefix notation (Section \ref{sec:results:boolean-calc-problem}).
We found that the regulation-on condition resulted in lower problem-solving success than the regulation-off condition. 
We also found that regulation-off solutions typically required fewer generations than regulation-on solutions to arise (Figure \ref{subfig:boolean-calc-postfix-solve-time}).
Additionally, we did not observe a significant difference in the proportion of flow-control instructions represented in execution traces of regulation-on and regulation-off solutions (supplement \supSecBooleanCalcPostfixAnalysis\ \citep{lalejini_supplement}). 

These results, in combination with our previous experimental results, suggest that tag-based regulation is beneficial when prior context dictates behavioral responses to input.
On such context-dependent problems, representations without explicit regulation must compensate with additional conditional logic structures. 

%representations lacking genetic regulation must compensate with larger conditional logic structures.

% @AML: Maybe one or two sentences putting this result (and maybe the changing signal problem result into a broader context?)

\section{Conclusion}

% -- bookmark --

% - Positive results, immediately applicable to other tag-enabled systems -
We demonstrated that tag-based genetic regulation allows GP systems to evolve programs with more dynamic plasticity.
These evolved programs are better able to solve context-dependent problems where the appropriate software modules to execute in response to a particular input changes over time.
Genetic regulation broadens the applicability of SignalGP, both as a representation for problem-solving and as a type of digital organism for studying evolutionary dynamics \citep{Lalejini_Moreno_Ofria_DISHTINY_2020}.
Further, this work illustrates an approach for easily incorporating tag-based models of gene regulation into existing GP systems. 

% - negative results, future work on matching schemes -
Our results also reveal that tag-based regulation is not necessarily beneficial across all problem domains.
We observed that the addition of tag-based regulation can impede adaptive evolution on problems where responses to inputs are not context-dependent (\textit{e.g.}, the independent-signal task and postfix version of the Boolean-logic calculator problem). 
A more thorough examination of what types of context-free problems are most sensitive to tag-based regulation---and how to mitigate any harm---would be potentially fruitful.

% ---- BEGIN REVISION ----
% - issues with existing representations -
Across all problems used in this work, the tag representation and matching scheme that we used was clearly sufficient for success. 
However, existing tag systems are limited in their capacity to scale up to substantially larger gene regulatory networks. 
As these networks grow, the specificity required for references to differentiate between modules increases. 
At some point references become brittle, as any mutation will switch the module that a call triggers.
In ongoing work, we are investigating the wide variations in scalability of different metrics for measuring the similarity between tags.
Substantial work will also need to be conducted by the community in order to develop more scalable representations for tag-based naming.
% @AML: new line below (currently very weak sentence)
For example, insights from the indirect referencing mechanisms of artificial biochemical networks and enzyme genetic programming systems may prove to be informative in developing new tag representations \citep{lones_artificial_2014,lones_biochemical_2013,lones_modelling_2004}.
% --- END REVISION ----

% ---- BEGIN REVISION ----
% - Comment: Another thing that would be good to comment on is the interpretability of the evolved solutions. I'm guessing that the regulatory-based solutions are harder for humans to understand?
% @AML: needs work!!
Evolved programs are often more challenging to read and understand than programs written by human developers.
In our experience, evolved programs that make use of tag-based regulation were substantially more difficult to read and interpret by hand than evolved programs that do not use tag-based regulation.
We found that visualizations of tag-based regulatory networks and program execution traces (\textit{e.g.}, Figures \ref{fig:signal-counting-example-networks} and \ref{fig:boolean-calc-prefix-example-networks}) greatly improved our ability to understand how a given evolved program worked.
As we scale up tag-based regulation, the development of interactive visualizations will become increasingly important for understanding evolved programs that use tag-based regulation.  
% ---- END REVISION -----

The current investigations have focused on regulation as a problem-solving tool, but with a few extensions these sorts of systems can also help us answer open questions about biological evolution.
Our current implementation of tag-based regulation facilitates plasticity only within a program's lifetime; if we extend this capacity across multiple generations, we can study the effects of epigenetic inheritance on evolutionary dynamics.
Epigenetic inheritance refers to heritable phenotypic changes that are not directly encoded by the underlying genetic sequence \citep{bender_plant_2002,jablonka_transgenerational_2009}.
For example, epigenetics is used in combination with gene regulation for cell-type differentiation in multicellular organisms \citep{mohn_genetics_2009, smith_dna_2013} and caste determination in some species of eusocial insects \citep{weiner_epigenetics_2012}.
SignalGP supports epigenetics with the addition of instructions that mark existing function regulation as heritable.
For our next steps, we will apply epigenetics-enabled SignalGP to study fraternal transitions in individuality and the evolution of differentiation before, during, and after a transition occurs \citep{Lalejini_Moreno_Ofria_DISHTINY_2020}.
Open-ended experiments with epigenetics and gene regulation will help illuminate the relationship between within-lifetime plastic adaptation and evolutionary adaptation over generational time scales.
Additionally, mechanisms for epigenetic inheritance have been shown to potentially improve GP performance \citep{la_cava_genetic_2015, la_cava_inheritable_2015,ricalde_evolving_2017}; as such, we plan to apply our insights back to automatic program synthesis.

\begin{acknowledgements}

We thank our anonymous reviewers and Clifford Bohm for feedback and suggestions on this manuscript. 
We also thank the Digital Evolution Laboratory for thoughtful discussions, ideas, and support.
This research was supported in part by NSF grants DEB-1655715, DBI-0939454, and  DGE-1424871, and by MSU through the computational resources provided by the Institute for Cyber-Enabled Research.
Any opinions, findings, and conclusions or recommendations expressed in this material are those of the authors and do not necessarily reflect the views of the NSF.
% The following lines are for the preprint!
This is a preprint of an article published in Genetic Programming and Evolvable Machines. The final
authenticated version is available online at: https://doi.org/10.1007/s10710-021-09406-8

% Reviewer thanks
% - important clarifications

\end{acknowledgements}

\section*{Conflict of interest}

The authors declare that they have no conflict of interest.

% \bibliographystyle{spmpsci}      % mathematics and physical sciences
\bibliographystyle{plain}
\small
\bibliography{bib/lit_references,bib/software_references}   % name your BibTeX data base

\end{document}